\documentclass[letterpaper]{article} 
\usepackage[submission]{aaai24}  
\usepackage{times}  
\usepackage{helvet}  
\usepackage{courier}  
\usepackage[hyphens]{url}  
\usepackage{graphicx} 
\usepackage{natbib}  
\usepackage{caption} 
\usepackage{algorithm}
\usepackage[noend]{algpseudocode}
\usepackage{amsmath}
\usepackage{lineno}
\usepackage{amssymb,bm}
\usepackage{booktabs}
\usepackage{diagbox}
\usepackage {threeparttable}
\usepackage{url}
\usepackage{xcolor}
\usepackage{multirow}
\usepackage{enumitem}
\usepackage{subfigure}
\usepackage[toc,title,page]{appendix}

\usepackage{newfloat}
\usepackage{listings}
\DeclareCaptionStyle{ruled}{labelfont=normalfont,labelsep=colon,strut=off} 
\floatstyle{ruled}
\newfloat{listing}{tb}{lst}{}
\floatname{listing}{Listing}
\title{TimeSQL: Improving Multivariate Time Series Forecasting with Multi-Scale Patching and Smooth Quadratic Loss}

\author{
    Site Mo\textsuperscript{\rm 1},
    Haoxin Wang\textsuperscript{\rm 1},
    Bixiong Li\textsuperscript{\rm 2},Songhai Fan\textsuperscript{\rm 3},Yuankai Wu\textsuperscript{\rm 4},Xianggen Liu\textsuperscript{\rm 4}\thanks{corresponding author\\
    liuxianggen@scu.edu.cn(Xianggen Liu)}
}
\affiliations{
    \textsuperscript{\rm 1}College of Electrical Engineering, Sichuan University, Chengdu 610065, China\\
    \textsuperscript{\rm 2}College of Architecture and Environment, Sichuan University, Chengdu 610065, China\\
    \textsuperscript{\rm 3}State Grid Sichuan Electric Power Research Institute, Chengdu 610041, China\\
    \textsuperscript{\rm 4}College of Computer Science, Sichuan University, Chengdu 610065, China
}

\begin{document}

\maketitle

\begin{abstract}
Time series is a special type of sequence data, a sequence of real-valued random variables collected at even intervals of time. The real-world multivariate time series comes with noises and contains complicated local and global temporal dynamics, making it difficult to forecast the future time series given the historical observations. This work proposes a simple and effective framework, coined as TimeSQL, which leverages multi-scale patching and smooth quadratic loss (SQL) to tackle the above challenges. The multi-scale patching transforms the time series into two-dimensional patches with different length scales, facilitating the perception of both locality and long-term correlations in time series. SQL is derived from the rational quadratic kernel and can dynamically adjust the gradients to avoid overfitting to the noises and outliers. Theoretical analysis demonstrates that, under mild conditions, the effect of the noises on the model with SQL is always smaller than that with MSE. Based on the two modules, TimeSQL achieves new state-of-the-art performance on the eight real-world benchmark datasets. Further ablation studies indicate that the key modules in TimeSQL could also enhance the results of other models for multivariate time series forecasting, standing as plug-and-play techniques. 
\end{abstract}

\section{Introduction}

Time series forecasting (TSF) aims to predict the time series values for a future period based on historical observations. It is a crucial task across diverse fields, including but not limited to power \cite{kardakos2013application}, transportation \cite{kadiyala2014multivariate}, and healthcare \cite{morid2023time}. However, this problem is very challenging. On the one hand, the modeling of locality and global correlations is difficult since the essential temporal dynamics underlying the time series are not observable. On the other hand, the real-world time series are real-valued and full of noises, hindering the machine-learning model from capturing the latent temporal dynamics.

\begin{figure*}[ht]
\centering
{
\includegraphics[width=0.7\textwidth]{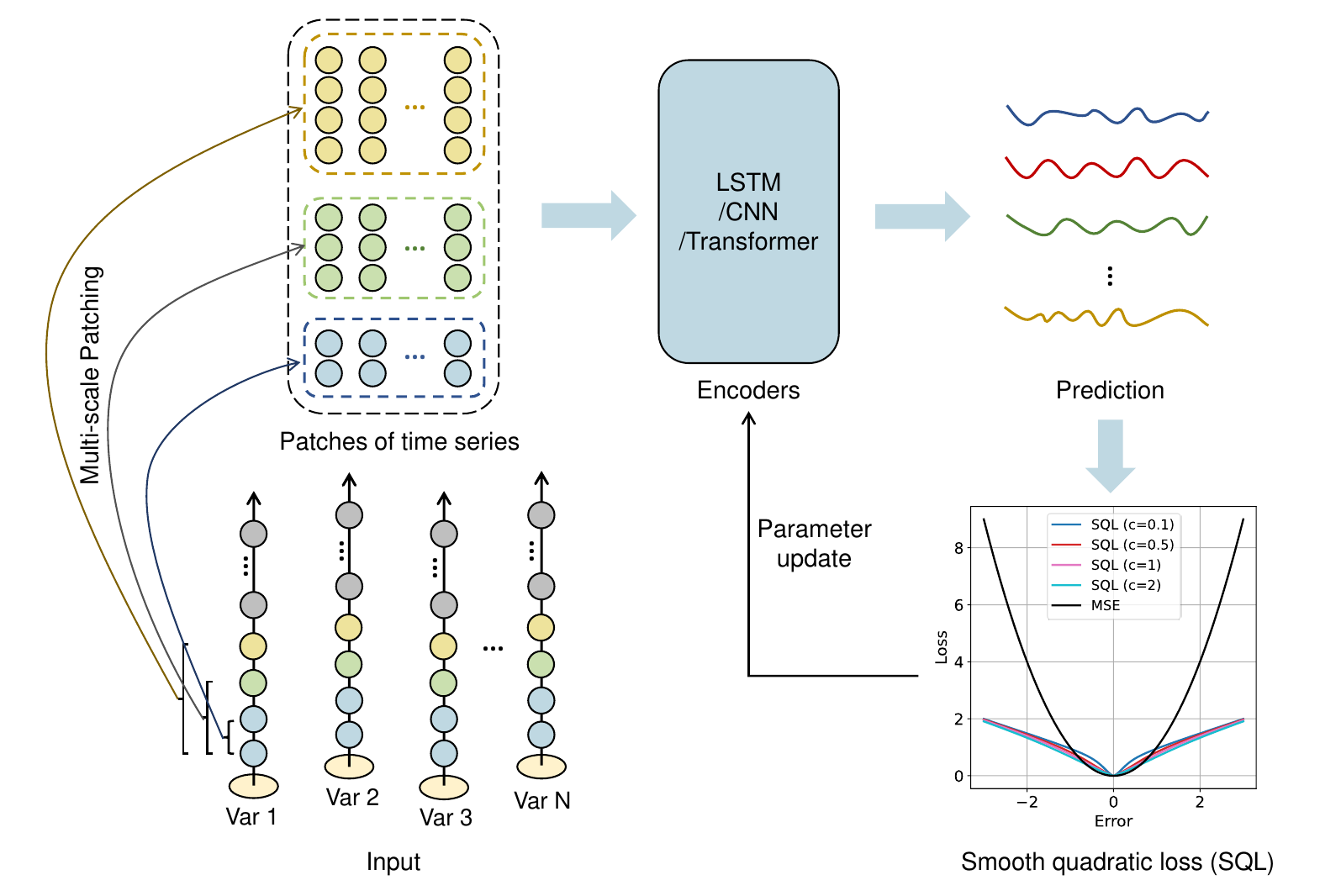}
\vspace{-3mm}
\caption{Overview of the TimeSQL framework.}\label{fig:Model overview}
\vspace{-3mm}
}
\end{figure*}
In the early years, TSF is typically accomplished by autoregressive integrated moving average (ARIMA) models \cite{bartholomew1971time}. To overcome the requirement for stationary and linear data of ARIMA, machine learning algorithms such as SVM \cite{hearst1998support} and XGBoost \cite{chen2015xgboost} gained attention for handling more complex time series in the real world. Recently, due to the impressive nonlinear feature extraction power, deep neural networks have dominated the advancements in TSF. Recurrent neural networks (e.g., GRU and LSTM) explicitly mimic the sequential modeling process in computations and are leveraged in TSF \cite{dey2017gate,graves2012long}. But they suffer the problem of unstable learning process \cite{zhu2019detecting} and catastrophic forgetting for long sequences \cite{mccloskey1989catastrophic}. 

In recent times, Transformer  \cite{vaswani2017attention} employs multi-head attention to effectively capture the
long-term dependencies in time series. The Transformer-based models, such as LogTrans \cite{li2019enhancing}, Pyraformer \cite{liu2021pyraformer}, Informer \cite{zhou2021informer}, Autoformer \cite{wu2021autoformer}, and FEDformer \cite{zhou2022fedformer}, PatchTST \cite{Yuqietal-2023-PatchTST}, have either enhanced computational efficiency or extraction capacities of the local and global properties. Most notably, PatchTST \cite{Yuqietal-2023-PatchTST} formulates the TSF into a regression problem on multiple sequence patches (akin to ``images''), achieving the current state-of-the-art forecasting results. 

Although Transformer-based methods make significant advancements in predicting the future time series, few of them consider eliminating the influences of the noises on the training process, which is also an inescapable challenge in TSF. In particular, the most widely used loss function, mean squared error (MSE) is sensitive to outliers since the MSE gradients are linear to the prediction error. With this objective function, the neural architectures inevitably fit the noises and even outliers in the time series. In the meanwhile, the other advanced loss functions such as approximated dynamic time warping \cite{cuturi2017soft} and DILATE \cite{le2019shape} concentrate on the sharp changes in non-stationary signals instead of  data noises. As a result, there is still a lack of the loss function that could reduce the influences of the noises on model learning.


In this work, we propose a novel smooth quadratic loss (SQL) function to guide the models to filter the noises and learn the underlying essential temporal law of the variable changing. The SQL function stems from 
a rational quadratic kernel and could dynamically adjust the gradient according to the prediction error, thus reducing the effects of the outliers. In addition, we also introduce a simple patching operation named multi-scale patching for efficient feature extraction. The multi-scale patching transforms the time series into two-dimensional patches with different scales, facilitating the perception of both locality and long-term correlations in time series. 

Based on the above two techniques, we build a simple and effective framework, coined as TimeSQL. Extensive experiments on 8 benchmark datasets show that TimeSQL outperforms all the other methods in most of the test settings, achieving new state-of-the-art TSF performance. In summary, the contributions of this work include:


\begin{itemize}
\item We propose smooth quadratic loss (SQL), for multivariate time series forecasting. It is theoretically proved that, under some mild conditions, the effect of the noises on the model with SQL is smaller than that with MSE.
\item We integrate SQL and the multi-scale patching operation to build TimeSQL. TimeSQL exhibits notably better results than the previous SOTA model, i.e., PatchTST.
\item Comprehensive ablation studies demonstrate the effectiveness and universality of the proposed SQL and  multi-scale patching. That is, they could also enhance the prediction capacities of the other TSF models.
\end{itemize}

\section{Preliminaries}
A data point in time series forecasting  contains historical data $X$ and the subsequent part of the time series $Y$ (also called ground truth). Time series forecasting aims to predict future time series based on historical data. Formally speaking, given the historical time series with window $L$: $\bm X=[\bm x^1,\dots, \bm x^L]\in \mathbb{R}^{N \times L}$, the model is required to predict future values with length $T$: $ \hat{\bm X} = [ \hat{\bm x}^{L+1},\dots, \hat{\bm x}^{L+T}]\in\mathbb{R}^{N \times T}$, where $\bm x^t \in \mathbb{R}^{N}$ is the values of the variables at $t^{th}$ step. $N$ stands for the number of the multivariate. The goal of a time series forecasting model is to minimize the difference between the prediction and the ground truth. 

It is noted that the number of variables (i.e., $N$) in the prediction equals the one in the input. According to the above notation, the superscript represents the index of time steps and the subscript denotes the index of the variable in the multivariate. For notation simplicity, we use $\bm x_n \in \mathbb{R}^{L}$ to indicate the time series of the $n^{th}$ variable. That is, the input $X$ could also be represented by $[\bm x_1,\dots, \bm x_N]^T$.

\section{Related Work}
With the continuous advancement of neural architectures, time series forecasting (TSF) research has achieved significant progression. Traditional recurrent neural networks (RNNs) \cite{medsker2001recurrent} are effective for sequential data but struggle with long sequences due to issues like gradient vanishing and exploding. The LSTM \cite{graves2012long}, GRU \cite{dey2017gate}, and LSTNet \cite{lai2018modeling} models have emerged as improved RNN-based solutions, demonstrating robustness in capturing long-term dependencies and excelling in TSF tasks. The temporal convolutional network \cite{bai2018empirical} introduces a CNN-based approach with multiple kernel sizes and outperforms canonical recurrent networks such as LSTMs.

Transformer models, by virtue of the ability of long-term feature extraction \cite{vaswani2017attention,radford2018improving}, have been extended to time series forecasting. LogTrans \cite{li2019enhancing} introduces convolutional self-attention with causal convolution to enhance the locality and reduce the memory requirement. Further, Informer \cite{zhou2021informer} improves the computational efficiency of the Transformer by dynamically selecting dominant queries in the attention mechanism. In addition, hierarchical pyramid attention in Pyraformer \cite{liu2021pyraformer}, seasonal decomposition in Autoformer \cite{wu2021autoformer}, and frequency-based attention in FEDFormer \cite{zhou2022fedformer} have also enhanced the temporal modeling of the Transformer architecture.

In addition, the neural architectures that are not Transformer based have also made non-trivial contributions in TSF. TimesNet \cite{wu2023timesnet} adopts a pure computer vision architecture to transform time series into 2D tensors. Similarly, MICN \cite{micn}  adopts CNN-based convolution for local features extraction and isometric convolution for global correlations discovery. In addition, \citeauthor{zeng2023transformers} \citeyear{zeng2023transformers} show that one-layer linear models could rival most of the Transformer-based models. 

Recently, PatchTST \cite{Yuqietal-2023-PatchTST} proposes to decompose the time series data into patches and then feed them into a ViT model \cite{dosovitskiy2020image}, standing as the current state-of-the-art model. Different from PatchTST, our work focuses on plug-and-play techniques for time series forecasting, i.e., multi-scale patching and smooth quadratic loss function, which are general and model-agnostic.




\section{Methodology}
In this section, we first introduce multi-scale patching to extract temporal dynamics and interactions within the time series. Then we elaborate on the smooth quadratic loss function to eliminate the distraction of the inevitable noises in time series.

\subsection{Multi-Scale Patching}
 The extraction of information at various scales is of paramount importance in time series prediction. Long-scale information facilitates the capture of broad patterns and changes over extended periods, which is highly valuable for forecasting long-term trends or cycles. Conversely, short-scale information enables the capture of fine-grained details and localized variations. By integrating information from different scales, predictive models can reap the benefits of a comprehensive understanding of the dynamics underlying the time series, resulting in more accurate predictions. To this end, we first propose a multi-scale patching operation to build the encoding of the time series. 
 


In time series, a single step of the variables could hardly have meaningful information in the temporal process. Therefore, inspired by \citeauthor{Yuqietal-2023-PatchTST} \citeyear{Yuqietal-2023-PatchTST}, we introduce a multi-scale patching operation to capture the locality of the semantic information in the time series. For a particular scale, the patching operation aggregates several time steps into a patch of the local semantic context of the time series. Formally speaking, the $i^{th}$ patch of the $n^{th}$ variable is denoted as $\bm p_{n,i}$, given by
\setlength{\abovedisplayskip}{4pt}
\setlength{\belowdisplayskip}{4pt}
\begin{align}
\!\!\!\!\!\!\bm p_{n,i}& = \bm X_{n,(i-1)*S^{(s)}:(i-1)*S^{(s)}+S^{(p)}} \\
&=[\bm x_n^{(i-1)*S^{(s)}},\cdots,x_n^{(i-1)*S^{(s)}+S^{(p)}}] \in \mathbb{R} ^{S^{(p)}}
\end{align}
where $S^{(p)}$ and $S^{(s)}$ stand for the patch scale and the sliding size, respectively. In total, there is $N^{(p)} = \lfloor\frac{L-S^{(p)}}{S^{(s)}} \rfloor+1$ patches, that is $i\in \{1,2,\cdots, N^{(p)}\}$. For example, for a single variable $\bm x_1$ with the length $L$, it could be transformed into $L-2$ patches if the patch scale $S^{(p)}$ is 3 and the sliding size $S^{(s)}$ is 1. Therefore, the patching operation transforms the input time series into multiple patches, represented as $\bm P$.
\begin{align}
\bm P &= \text{Patching}(\bm X,S^{(p)},S^{(s)})  \in \mathbb{R}^{N \times N^{(p)}\times S^{(p)}}
\end{align}
In this way, the time series is divided into multiple local contexts, similar to the discrete words in the natural language.

The above patching operation captures the semantic contexts of time series with a single scale. Next, we adopt $K$ patching scales and patching striding sizes to capture sufficient information underlying the temporal dynamics, which is coined as multi-scale patching. 
\begin{align}
\bm P^{(k)} &= \text{Patching}(\bm X,S^{(p,k)},S^{(s,k)}),k=1,\cdots, K 
\end{align}
where $\bm P^{(k)}$ stands for the patching features extracted by the $k$-th patching scale $S^{(p,k)}$ and patching sliding size $S^{(s,k)}$.



\subsection{Multi-Scale Feature Integration} 
As the patches in individual groups capture distinct temporal characteristics, we leverage $K$ temporal encoders to learn the corresponding representations independently, given by
\begin{align}
\bm  h_{k}=\text{Encoder}_{k}(\bm P^{(k)}) \in \mathbb{R}^{N\times N^{(p)}  \times H} ,
\end{align}
where $\bm  h_{k}$ denotes the output features of the $k^{th}$ temporal encoders. $H$ is the hidden size of the encoder. As the sizes of the multi-scale patches are different, the input sizes of the temporal encoders are adapted to fit the patch sizes. As TimeSQL is a general framework, the temporal encoders could be implemented by LSTM, Transformer, CNN, etc.

The representations processed by the temporal encoders are further concatenated together as the features of the historical time series, given by $\bm  = [\bm h_{1},\bm h_{2},\cdots,\bm h_{K}]$. Finally, we employ a multi-layer perceptron (MLP) to predict the future values of the time series.
\begin{equation}
    \begin{aligned}
 \hat{X}=\text{MLP}([\bm h_{1},\bm h_{2},\cdots,\bm h_{K}])  \\
    \end{aligned}
\end{equation}
Where $\hat{X}$ is the predicted time series of TimeSQL.
\subsection{Smooth Quadratic Loss}
Due to the unobservable hidden states of the time series, the noises are inevitable in temporal dynamics. Consequently, one of the long-standing problems in time series is how to filter the noises and learn the underlying essential temporal law of the variable changing. The traditional loss functions, such as L1 and L2, do not consider the existence of noises in the ground truth. For example, the gradients under mean square error are linearly increasing upon the prediction error. In this work, we propose a loss function, named smooth quadratic loss, that could dynamically adjust the gradient according to the prediction error. The smooth quadratic loss is designed based on the rational quadratic kernel function, which calculates the difference between the prediction and the ground truth in a high-dimensional space. 


\begin{figure*}[ht]
\centering
\includegraphics[width=0.96\textwidth]{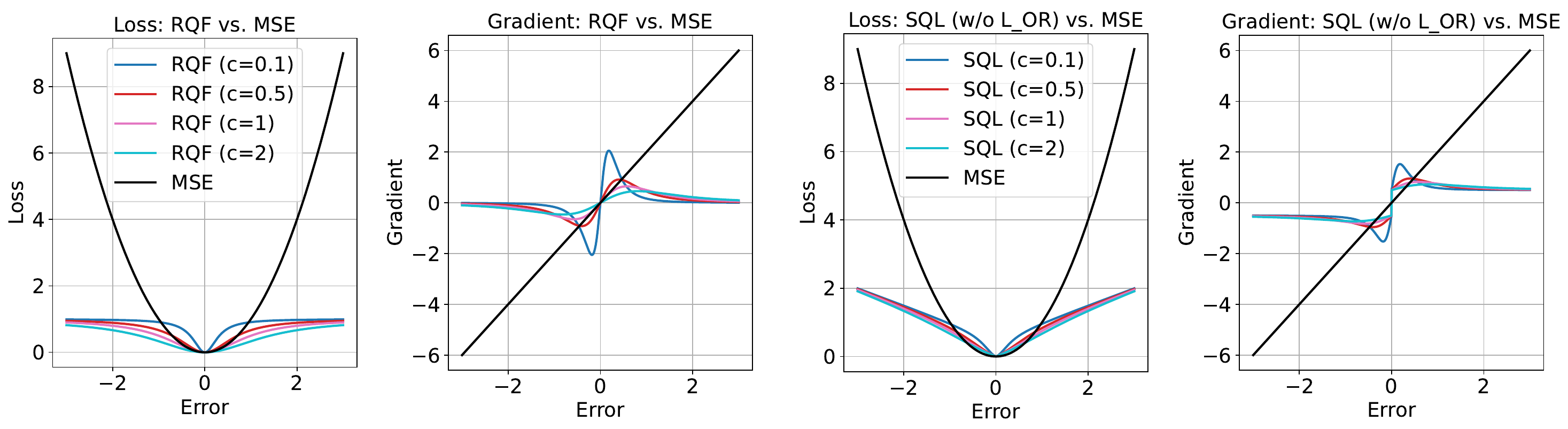}\vspace{-3mm}
\caption{The curves of the individual loss functions and the corresponding gradients.}\label{fig:RQF}\vspace{-3mm}
\end{figure*}

\noindent\textbf{Rational Quadratic Function.} Assuming $\Phi$ is a nonlinear mapping function from the original feature space to the high-dimensional feature space. The inner product in the kernel space can be defined as follows:
\begin{align}
\kappa ({\bm u},{\bm v})=<\Phi({\bm u}),\Phi({\bm v})>,
\end{align}
where ${\bm u}$ and ${\bm v}$ are two vectors.

In this study, we propose the use of the rational quadratic function (RQF) as an alternative to the commonly used Gaussian kernel function. The specific form of the rational quadratic function is given by:
\begin{equation}
    \begin{aligned}
        & \kappa\left({\bm u}, {\bm v}\right)=1-\frac{({\bm u}-{\bm v})^2}{({\bm u}-{\bm v})^2+ c},  \\
    \end{aligned}
\end{equation}
where $c$ is a hyperparameter. Therefore,
the RQF loss for a single data point in time series forecasting is:
\begin{equation}
    \begin{aligned}
        & L_{RQF}(\hat{ x},\hat{ y})=1-\kappa\left(\hat{ x}, \hat{ y}\right)
        = \frac{(\hat{ x}-\hat{ y})^2}{(\hat{x}-\hat{ y})^2+c},  \\
    \end{aligned}
\end{equation}
where $\hat{x}\in \mathbb{R}$ and $\hat{y}\in \mathbb{R}$ stand for the prediction and ground truth of the single data point. Besides its complex non-linearity, this loss function is distinguished by its broad scope and rapid computation but is known to be sensitive to the choice of parameters.

The gradient of the rational quadratic loss respective to the prediction $\hat{\bm x}$ is given by: 
\begin{align}
  \frac{\partial L_{RQF}}{\partial \hat{ x}} &= \frac{\partial L_{RQF}}{\partial  e} = \frac{2 c  e}{({ e}^2 +  c)^2},
\end{align}
where $ e$ stands for the prediction error, i.e., $ e = {\hat{ x} - \hat{ y}}$. 

Intriguingly, we observe that the gradient of the rational quadratic function (RQF) loss is nonlinear to the prediction error and insensitive to the outliers in the time series (Figure~\ref{fig:RQF}). When the scale of prediction error progressively increases (one possible reason is the growing noises), the amplitude of the RQF gradient is first increasing and then gradually decreasing to a certain constant. This characteristic of the RQF gradient enforces the model first to optimize the prediction error of normal data points. For abnormal, difficult data points or even outliers, it adopts a smaller strength in optimization. As for MSE, its gradient regarding the prediction error is linear, which is not agile enough for processing the noising time series.

\noindent \textbf{Theoretical Analysis.} We provide theoretical justifications for the proposed rational quadratic loss function. Below, we investigate how the noises in the ground truth influence the learning of the time series forecasting models.
\newtheorem{definition}{Definition}
\begin{definition}
Let $ \varepsilon$ and $ y$ be the noise and the noiseless ground truth in the label $\hat{ y}$ (i.e., $\hat{ y}=  y + \varepsilon$). The effect of the noise on the model is evaluated by the normalized derivation of the loss function, calculated by $|\frac{ f(y+ \varepsilon,\hat{ x})-  f( y,\hat{ x})}{f( y,\hat{ x})}|$, where  $\hat{\bm x}$ stand for the model prediction, respectively.
\end{definition}
\newtheorem{theorem}{Theorem}
\begin{theorem}
Regardless of the distribution of the noise $\varepsilon$ in the time-series data, the effect of the noise on the RQF loss is always lower than that of MSE loss.
\end{theorem}
The proof is in Appendix A.
\begin{theorem}
If $|\varepsilon|\ge 2|\hat{x}-y|$, we have
\begin{align}
&\!\!\!\!\!\!\!\!\!\!\!\left | \frac{\nabla_\theta RQF(y+\varepsilon,\hat{x})- \nabla_\theta RQF(y,\hat{x})}{\nabla_\theta  RQF(y,\hat{x})}\right | \nonumber \\  & \le \left | \frac{\nabla_\theta MSE(y+\varepsilon,\hat{x})  - \nabla_\theta MSE(y,\hat{x})}{\nabla_\theta MSE(y,\hat{x})}\right |
\end{align}
\end{theorem}
Indeed, a tighter constraint exists for the potential range of the noise $\varepsilon$ (partially dependent on $c$). However, since an analytic solution is unavailable, we resort to employing mathematical relaxations to make approximations. The detailed proof sketch is in Appendix B.

\begin{table*}[!h]
	\centering
    \renewcommand{\arraystretch}{1.15}
	\resizebox{\linewidth}{!}{
		\begin{tabular}{cc|c|ccccccccccccccccccc} \toprule
			&\multicolumn{2}{c}{Models}& \multicolumn{2}{c}{TimeSQL}& \multicolumn{2}{c}{PatchTST}& \multicolumn{2}{c}{DLinear}& \multicolumn{2}{c}{MICN}& \multicolumn{2}{c}{FEDformer}& \multicolumn{2}{c}{Autoformer}& \multicolumn{2}{c}{Informer}& \multicolumn{2}{c}{Pyraformer}& \multicolumn{2}{c}{LogTrans}& \\
			\cmidrule(r){4-5}\cmidrule(r){6-7}\cmidrule(r){8-9}\cmidrule(r){10-11}\cmidrule(r){12-13}\cmidrule(r){14-15}\cmidrule(r){16-17}\cmidrule(r){18-19}\cmidrule(r){20-21}
			&\multicolumn{2}{c|}{Metric}&MSE&MAE&MSE&MAE&MSE&MAE&MSE&MAE&MSE&MAE&MSE&MAE&MSE&MAE&MSE&MAE&MSE&MAE\\ \midrule
			&\multirow{4}*{{Weat.}}& 96    & \textbf{0.148} & \textbf{0.185} & 0.152 & 0.199 & 0.176 & 0.237 & 0.173 & 0.239 & 0.238 & 0.314 & 0.249 & 0.329 & 0.354 & 0.405 & 0.896 & 0.556 & 0.458 & 0.490 \\
            &\multicolumn{1}{c|}{}& 192   & \textbf{0.190} & \textbf{0.227} & 0.197 & 0.243 & 0.220 & 0.282 & 0.220 & 0.287 & 0.275 & 0.329 & 0.325 & 0.370 & 0.419 & 0.434 & 0.622 & 0.624 & 0.658 & 0.589 \\
            &\multicolumn{1}{c|}{}& 336   & \textbf{0.243} & \textbf{0.268} & 0.249 & 0.283 & 0.265 & 0.319 & 0.277 & 0.331 & 0.339 & 0.377 & 0.351 & 0.391 & 0.583 & 0.543  & 0.739 & 0.753 & 0.797 & 0.652 \\
            &\multicolumn{1}{c|}{}& 720   & 0.322 & \textbf{0.323} & \textbf{0.320} & 0.335 & 0.323 & 0.362 & 0.316 & 0.355 & 0.389 & 0.409 & 0.415 & 0.426 & 0.916 & 0.705 & 1.004 & 0.934 & 0.869 & 0.675 \\  \midrule
			&\multirow{4}*{{Traf.}}& 96    & 0.380 & \textbf{0.234} & \textbf{0.367} & 0.251 & 0.410 & 0.282 & 0.456 & 0.288 & 0.576 & 0.359 & 0.597 & 0.371 & 0.733 & 0.410 & 2.085 & 0.468 & 0.684 & 0.384 \\
            &\multicolumn{1}{c|}{} & 192   & 0.400 & \textbf{0.242} & \textbf{0.385} & 0.259 & 0.423 & 0.287 & 0.486 & 0.299 & 0.610 & 0.380 & 0.607 & 0.382 & 0.777 & 0.435 & 0.867 & 0.467 & 0.685 & 0.390 \\
            &\multicolumn{1}{c|}{}& 336   & 0.412 & \textbf{0.249} & \textbf{0.398} & 0.265 & 0.436 & 0.296 & 0.491 & 0.303 & 0.608 & 0.375 & 0.623 & 0.387 & 0.776 & 0.434 & 0.869 & 0.469 & 0.734 & 0.408 \\
            &\multicolumn{1}{c|}{}& 720   & 0.444 & \textbf{0.269} & \textbf{0.434} & 0.287 & 0.466 & 0.315 & 0.525 & 0.355 & 0.621 & 0.375 & 0.639 & 0.395 & 0.827 & 0.466 & 0.881 & 0.473 & 0.717 & 0.396 \\  \midrule
			&\multirow{4}*{{Elec.}}& 96    & \textbf{0.130} & \textbf{0.219} & \textbf{0.130} & 0.222 & 0.140 & 0.237 & 0.155 & 0.264 & 0.186 & 0.302 & 0.196 & 0.313 & 0.304 & 0.393 & 0.386 & 0.449 & 0.258 & 0.357 \\
			&\multicolumn{1}{c|}{}& 192   & \textbf{0.147} & \textbf{0.235} & 0.148 & 0.240 & 0.153 & 0.249& 0.167 & 0.276 & 0.197 & 0.311 & 0.211 & 0.324 & 0.327 & 0.417 & 0.386 & 0.443 & 0.266 & 0.368 \\
			&\multicolumn{1}{c|}{}& 336   & \textbf{0.164} & \textbf{0.253} & 0.167 & 0.261 & 0.169 & 0.267 & 0.199 & 0.307 & 0.213 & 0.328 & 0.214 & 0.327 & 0.333 & 0.422 & 0.378 & 0.443 & 0.280 & 0.380 \\
			&\multicolumn{1}{c|}{}& 720   & 0.204 & \textbf{0.287} & \textbf{0.202} & 0.291 & 0.203 & 0.301 & 0.214 & 0.323 & 0.233 & 0.344 & 0.236 & 0.342 & 0.351 & 0.427 & 0.376 & 0.445 & 0.283 & 0.376 \\  \midrule
			&\multirow{4}*{{ILI}}& 24    & \textbf{1.298} & \textbf{0.665} & 1.522 & 0.814 & 2.215 & 1.081 & 2.416 & 1.051 & 2.624 & 1.095 & 2.906 & 1.182 & 4.657 & 1.449  & 1.420 & 2.012 & 4.480 & 1.444 \\
            &\multicolumn{1}{c|}{} & 36    & \textbf{1.241} & \textbf{0.676} & 1.430 & 0.834 & 1.963 & 0.963 & 2.265 & 0.988 & 2.516 & 1.021 & 2.585 & 1.038 & 4.650 & 1.463 & 7.394 & 2.031 & 4.799 & 1.467 \\
            &\multicolumn{1}{c|}{}& 48    & \textbf{1.530} & \textbf{0.750} & 1.673 & 0.854 & 2.130 & 1.024 & 2.296 & 1.037 & 2.505 & 1.041 & 3.024 & 1.145 & 5.004 & 1.542 & 7.551 & 2.057 & 4.800 & 1.468 \\
            &\multicolumn{1}{c|}{}& 60    & \textbf{1.406} & \textbf{0.731} & 1.529 & 0.862 & 2.368 & 1.096 & 2.751 & 1.173 & 2.742 & 1.122 & 2.761 & 1.114 & 5.071 & 1.543 & 7.662 & 2.100 & 5.278 & 1.560 \\  \midrule
			&\multirow{4}*{{ETTh1}}& 96    & \textbf{0.360} & \textbf{0.386} & 0.375 & 0.399 & 0.375 & 0.399 & 0.408 & 0.432 & 0.376 & 0.415 & 0.435 & 0.446 & 0.941 & 0.769 & 0.664 & 0.612 & 0.878 & 0.740 \\
            &\multicolumn{1}{c|}{}& 192   & \textbf{0.402} & \textbf{0.412} & 0.414 & 0.421 & 0.405 & 0.416 & 0.453 & 0.472 & 0.423 & 0.446 & 0.456 & 0.457 & 1.007 & 0.786 & 0.790 & 0.681 & 1.037 & 0.824 \\
            &\multicolumn{1}{c|}{}& 336   & \textbf{0.414} & \textbf{0.421} & 0.431 & 0.436 & 0.439 & 0.443 & 0.575 & 0.549 & 0.444 & 0.462 & 0.486 & 0.487 & 1.038 & 0.784 & 0.891 & 0.738 & 1.238 & 0.932 \\
            &\multicolumn{1}{c|}{}& 720   & \textbf{0.420} & \textbf{0.446} & 0.449 & 0.466 & 0.472 & 0.490 & 0.716 & 0.645 & 0.469 & 0.492 & 0.515 & 0.517 & 1.144 & 0.857 & 0.963 & 0.782 & 1.135 & 0.852 \\  \midrule
			&\multirow{4}*{{ETTh2}}& 96    & \textbf{0.274} & \textbf{0.328} & \textbf{0.274} & 0.336 & 0.289 & 0.353 & 0.287 & 0.352 & 0.332 & 0.374 & 0.332 & 0.368 & 1.549 & 0.952 & 0.645 & 0.597 & 2.116 & 1.197 \\
            &\multicolumn{1}{c|}{}& 192   & \textbf{0.339} & \textbf{0.371} & \textbf{0.339} & 0.379 & 0.383 & 0.418 & 0.377 & 0.413 & 0.407 & 0.446 & 0.426 & 0.434 & 3.792 & 1.542 & 0.788 & 0.683 & 4.315 & 1.635 \\
            &\multicolumn{1}{c|}{}& 336   & \textbf{0.330} & \textbf{0.373} & 0.331 & 0.380 & 0.448 & 0.465 & 0.687 & 0.597 & 0.400 & 0.447 & 0.477 & 0.479 & 4.215 & 1.642 & 0.907 & 0.747 & 1.124 & 1.604 \\
            &\multicolumn{1}{c|}{}& 720   & 0.382 & \textbf{0.415} & \textbf{0.379} & 0.422 & 0.605 & 0.551 & 1.173 & 0.801 & 0.412 & 0.469 & 0.453 & 0.490 & 3.656 & 1.619 & 0.963 & 0.783 & 3.188 & 1.540 \\  \midrule
			&\multirow{4}*{{ETTm1}}& 96    & \textbf{0.283} & \textbf{0.328} & 0.290 & 0.342 & 0.299 & 0.343 & 0.298 & 0.349 & 0.326 & 0.390 & 0.510 & 0.492 & 0.626 & 0.560 & 0.543 & 0.510 & 0.600 & 0.546 \\
            &\multicolumn{1}{c|}{}& 192   & \textbf{0.324} & \textbf{0.355} & 0.332 & 0.369 & 0.335 & 0.365 & 0.343 & 0.382 & 0.365 & 0.415 & 0.514 & 0.495 & 0.725 & 0.619 & 0.557 & 0.537 & 0.837 & 0.700 \\
            &\multicolumn{1}{c|}{}& 336   & \textbf{0.356} & \textbf{0.376} & 0.366 & 0.392 & 0.369 & 0.386 & 0.400 & 0.419 & 0.392 & 0.425 & 0.510 & 0.492 & 1.005 & 0.741 & 0.754 & 0.655 & 1.124 & 0.832 \\
            &\multicolumn{1}{c|}{}& 720   & 0.424 & \textbf{0.420} & \textbf{0.416} & \textbf{0.420} & 0.424 & 0.421 & 0.529 & 0.500 & 0.446 & 0.458 & 0.527 & 0.493 & 1.133 & 0.845 & 0.908 & 0.724 & 1.153 & 0.820 \\  \midrule
			&\multirow{4}*{{ETTm2}} & 96    & \textbf{0.163} & \textbf{0.246} & 0.165 & 0.255 & 0.167 & 0.260 & 0.174 & 0.272 & 0.180 & 0.271 & 0.205 & 0.293 & 0.355 & 0.462& 0.435 & 0.507 & 0.768 & 0.642 \\
            &\multicolumn{1}{c|}{}& 192   & \textbf{0.216} & \textbf{0.283} & 0.220 & 0.292 & 0.224 & 0.303 & 0.240 & 0.320 & 0.252 & 0.318 & 0.278 & 0.336 & 0.595 & 0.586 & 0.730 & 0.673 & 0.989 & 0.757 \\
            &\multicolumn{1}{c|}{}& 336   & \textbf{0.264} & \textbf{0.315} & 0.278 & 0.329 & 0.281 & 0.342 & 0.353 & 0.404 & 0.324 & 0.364 & 0.343 & 0.379 & 1.270 & 0.871 & 1.201 & 0.845 & 1.334 & 0.872 \\
            &\multicolumn{1}{c|}{}& 720   & \textbf{0.348} & \textbf{0.370} & 0.367 & 0.385 & 0.397 & 0.421 & 0.438 & 0.444 & 0.410 & 0.420 & 0.414 & 0.419 & 3.001 & 1.267 & 3.625 & 1.451 & 3.048 & 1.328 \\  \midrule
            &\multicolumn{2}{c|}{Average}   & \textbf{0.436} & \textbf{0.364} & 0.460 & 0.391 & 0.562 & 0.437 & 0.652 & 0.476 & 0.651 & 0.472 & 0.713 & 0.497 & 1.629 & 0.825 & 1.528 & 0.820 & 1.592 & 0.860 \\ \bottomrule
		\end{tabular}
	}\vspace{-1mm}
	\caption{Multivariate long-term forecasting results on eight benchmark datasets. The best results are in \textbf{bold} .}
 \vspace{-2mm}
	\label{tab:multivariate}
\end{table*}

\noindent\textbf{Remark 1}. Theorems 1 and 2 demonstrate that when the ground truth is mixed with noises, both the loss values of the rational quadratic function (RQF) and the corresponding gradients align better with the ground truth than the MSE loss. In other words, the effect of the noises on the RQF loss is always smaller than the MSE loss. Since the above claim is supported in terms of both the loss value and the gradient, we can conclude that the effect of the noises on the learning process will be also smaller if the RQF loss is used.

\begin{table}[t]
\centering
\resizebox{\linewidth}{!}{
\begin{tabular}{c|cccccccc}
\toprule  
Datasets & Weat.  & Traf. & Elec. & ILI & ETTh1 & ETTh2 & ETTm1 & ETTm2 \\
\midrule  
 \# features & 21 & 862 & 321 & 7 & 7 & 7 & 7 & 7 \\
\# samples & 52696 & 17544 & 26304 & 966 & 17420 & 17420 & 69680 & 69680 \\
Interval & 10 mins & 1 hour & 1 hour & 1 week & 1 hour & 1 hour & 15 mins & 15 mins \\
\bottomrule 
\end{tabular}
}\vspace{-1mm}
\caption{Statistics of the benchmark  datasets.}
\vspace{-4mm}
\label{tab:data}
\end{table}

\noindent\textbf{Smooth Quadratic Loss}. We apply the McLaughlin formula (i.e., a particular form of the Taylor series) \cite{maclaurin1742treatise}  to further investigate the characteristics of RQF. RQF is approximated by  
\begin{equation}
    \begin{aligned}
          L_{RQF}(\hat{x},\hat{y})=\sum_{i=1}^{n} (-1)^{i-1}\frac{(\hat{x}-\hat{y})^{2i}}{c^i} +o\left((\hat{x}-\hat{y})^{2 n}\right) .
    \end{aligned}
\end{equation}
Notably, RQF is highly nonlinear to measure the prediction errors. In the meanwhile, we observe that the lowest-order term of RQF is quadratic. RQF may reckon without the low-latency information in time series. Thus, we integrate the mean absolute error (MAE) function into RQF to make the loss function smoother.   
\begin{equation}
    \begin{aligned}
         &L_{SQL}=\frac{\alpha}{T}\sum_{t=1}^{T} \frac{(\hat{x}_t-\hat{y}_t)^2}{(\hat{x}_t-\hat{y}_t)^2+c} +\frac{1-\alpha}{T}\sum_{t=1}^{T}\left | \hat{x}_t-\hat{y}_t \right | , \\
    \end{aligned}
    \label{rqf+mae}
\end{equation}
where $\alpha$ is a smooth coefficient.

As another technique to avoid overfitting to the outliers with large amplitudes, an outlier regularization (OR) on the predicted time series is proposed. Specifically, we apply the L1 and L2 regularization to the predicted value of the TimeSQL model with small scaling coefficients separately. In this way, OR penalizes the predicted time series that is substantially away from 0, encouraging the activations to remain small and smooth. While most of the regularization techniques are applied to the learnable weights or the layer-wise activations (e.g., layer normalization), the outlier regularization directly regularizes the output, which is new and has shown effectiveness in our experiments.

In summary, the SQL (smooth quadratic loss) is the combination of RQF, MAE, and outlier regularization (OR):
\begin{align}
L_{SQL}&=\alpha L_{RQF}+(1-\alpha )L_{MAE}+L_{OR} \nonumber \\
& =\frac{1}{T} \sum_{t=1}^{T} \Big [\alpha \frac{(\hat{x}_t-\hat{y}_t)^2}{(\hat{x}_t-\hat{y}_t)^2+c}+(1-\alpha)\left | \hat{x}_t- \hat{y}_t \right | \nonumber \\
        &+\beta\left | \hat{x}_{t}  \right | +\gamma \hat{x}_{t}  ^2 \Big],
\end{align}
where $\beta$ and $\gamma$ are the hyperparameters.

\section{Experiments}
\subsection{Datasets and Competitive Methods}
We used 8 publicly available datasets commonly used for time series forecasting to evaluate our method, covering various fields, including Weather, Traffic, Electricity, ILI, and ETT (ETTh1, ETTh2, ETTm1, ETTm2) \cite{zhou2021informer,wu2021autoformer}. Most of the time series in these datasets are long-term, requiring the model to capture the historical dynamics. The details of each dataset are shown in Table~\ref{tab:data}.

We have selected various state-of-the-art (SOTA) models that have emerged in recent years, including multi-scale isometric convolution network (MICN) \cite{micn}, LogTrans \cite{li2019enhancing}, Pyraformer \cite{liu2021pyraformer}, Informer \cite{zhou2021informer}, AutoFormer \cite{wu2021autoformer}, decomposition linear model (Dlinear) \cite{zeng2023transformers},  frequency enhanced decomposed Transformer (FEDFormer) \cite{zhou2022fedformer}, and channel-independent patch time series Transformer (PatchTST) \cite{Yuqietal-2023-PatchTST}. In particular, Dlinear and PatchTST are the previously best MLP-based and Transformer-based methods, respectively. As for the comparison of the loss functions, although the approximated dynamic time warping \cite{cuturi2017soft} and DILATE \cite{le2019shape} functions are designed for time series forecasting, they mainly focus on the sharp changes in non-stationary signals instead of noises. Also, they are only applicable to the simple time series with only one variable, which is not comparable to TimeSQL.

\begin{table}[t]
 \centering
 \renewcommand{\arraystretch}{1.15}
  \resizebox{1.\linewidth}{!}{
		\begin{tabular}{cc|c|cccc|ccccc}
			\toprule
			&\multicolumn{2}{c}{Methods}& \multicolumn{2}{c}{TimeSQL}& \multicolumn{2}{c}{TimeSQL+SP}& \multicolumn{2}{c}{TimeSQL+MPT}& \multicolumn{2}{c}{TimeSQL+SPT} \\
			\cmidrule(r){4-5}\cmidrule(r){6-7}\cmidrule(r){8-9}\cmidrule(r){10-11}
			&\multicolumn{2}{c|}{Metric}&MSE&MAE&MSE&\multicolumn{1}{c}{MAE}&MSE&MAE&MSE&MAE\\
			\midrule
			&\multirow{4}*{{ILI}}& 24    & \textbf{1.298} & \textbf{0.665} & 1.341 & 0.678 & \textbf{1.392} & \textbf{0.695} & 1.412 & 0.722\\
            &\multicolumn{1}{c|}{}& 36    & \textbf{1.241} & \textbf{0.676} & 1.273 & 0.688 & \textbf{1.349} & \textbf{0.718} & 1.351 & 0.714\\
            &\multicolumn{1}{c|}{}& 48    & 1.530 & 0.750 & \textbf{1.504} & \textbf{0.749} & \textbf{1.582} & \textbf{0.772} & 1.634 & 0.789\\
            &\multicolumn{1}{c|}{}& 60    & \textbf{1.406} & \textbf{0.731} & 1.444 & 0.736 & 1.460 & 0.753 & 1.453 & 0.755\\
			\midrule
			&\multirow{4}*{{ETTh1}}& 96    & \textbf{0.360} & \textbf{0.386} & 0.375 & 0.395 & \textbf{0.360} & \textbf{0.386} & \textbf{0.360} & 0.387\\
            &\multicolumn{1}{c|}{}& 192   & \textbf{0.402} & \textbf{0.412} & 0.418 & 0.419 & \textbf{0.407} & \textbf{0.417} & 0.407 & 0.418\\
            &\multicolumn{1}{c|}{}& 336   & \textbf{0.414} & \textbf{0.421} & 0.420 & 0.424 & \textbf{0.425} & 0.427 & \textbf{0.422} & 0.426\\
            &\multicolumn{1}{c|}{}& 720   & \textbf{0.420} & \textbf{0.446} & 0.429 & 0.450 & \textbf{0.420} & \textbf{0.446} & 0.427 & 0.452\\
			\midrule
			&\multirow{4}*{{ETTh2}}& 96    & \textbf{0.274} & \textbf{0.328} & 0.294 & 0.342 & 0.274 & \textbf{0.329} & \textbf{0.273} & \textbf{0.329}\\
            &\multicolumn{1}{c|}{}& 192   & \textbf{0.339} & \textbf{0.371} & 0.357 & 0.384 & 0.344 & 0.374 & \textbf{0.341} & \textbf{0.373}\\
            &\multicolumn{1}{c|}{}& 336   & \textbf{0.330} & \textbf{0.373} & 0.335 & 0.384 & 0.336 & 0.378 & \textbf{0.331} & \textbf{0.376}\\
            &\multicolumn{1}{c|}{}& 720   & \textbf{0.382} & \textbf{0.415} & 0.403 & 0.432 & 0.386 & 0.422 & \textbf{0.378} & \textbf{0.415}\\
			\midrule
			&\multirow{4}*{{ETTm1}}& 96    & \textbf{0.283} & \textbf{0.328} & 0.289 & 0.329 & \textbf{0.280} & \textbf{0.325} & 0.287 & 0.329\\
            &\multicolumn{1}{c|}{}& 192   & \textbf{0.324} & \textbf{0.355} & 0.334 & 0.356 & \textbf{0.331} & \textbf{0.356} & \textbf{0.331} & 0.357\\
            &\multicolumn{1}{c|}{}& 336   & \textbf{0.356} & \textbf{0.376} & 0.368 & 0.378 & \textbf{0.365} & \textbf{0.377} & 0.371 & 0.380\\
            &\multicolumn{1}{c|}{}& 720   & \textbf{0.424} & 0.420 & \textbf{0.424} & \textbf{0.414} & \textbf{0.419} & \textbf{0.412} & 0.424 & 0.414\\
			\midrule
			&\multirow{4}*{{ETTm2}} & 96    & 0.163 & \textbf{0.246} & \textbf{0.162} & \textbf{0.246} &\textbf{0.160} & \textbf{0.243} & 0.161 & 0.246\\
            &\multicolumn{1}{c|}{}& 192   & \textbf{0.216} & \textbf{0.283} & \textbf{0.216} & 0.284 & \textbf{0.216} & \textbf{0.283} & 0.219 & 0.286\\
            &\multicolumn{1}{c|}{}& 336   & \textbf{0.264} & \textbf{0.315} & 0.266 & 0.317 & \textbf{0.264} & \textbf{0.316} & 0.269 & 0.320\\
            &\multicolumn{1}{c|}{}& 720   & \textbf{0.348} & \textbf{0.370} & 0.351 & \textbf{0.370} & 0.350 & 0.372 & \textbf{0.348} & \textbf{0.371}\\
			\midrule
            &\multicolumn{2}{c|}{Average} & \textbf{0.539} & \textbf{0.433} & 0.550 & 0.439 & \textbf{0.556} & \textbf{0.440} & 0.561 & 0.443\\
			\bottomrule
		\end{tabular}
	}\vspace{-1mm}
\caption{The performance comparisons between multi-scale patching (MP) and single-scale patching (SP). The character ``T'' in MPT and SPT stands for the temporal encoder in TimeSQL is implemented by Transformer.}\vspace{-2mm}
\label{tab:TimeSQL}
\end{table}

\begin{table}[t]
    \centering
    \renewcommand{\arraystretch}{1.15}
    \resizebox{0.98\linewidth}{!}{
    \begin{tabular}{cc|c|ccccccccccc}
        \toprule
        & \multicolumn{2}{c}{Loss Functions} & \multicolumn{2}{c}{TimeSQL} & \multicolumn{2}{c}{TimeSQL-RQF} & \multicolumn{2}{c}{TimeSQL-OR} & \multicolumn{2}{c}{TimeSQL-MAE} & \multicolumn{2}{c}{TimeSQL-SQL} \\
        \cmidrule(r){4-5}\cmidrule(r){6-7}\cmidrule(r){8-9}\cmidrule(r){10-11}\cmidrule(r){12-13}
        & \multicolumn{2}{c|}{Metric} & MSE & MAE & MSE & MAE & MSE & MAE & MSE & MAE & MSE & MAE \\
        \midrule
        & \multirow{4}{*}{{ILI}} & 24 & 1.298 & \textbf{0.665} & 1.298 & 0.666 & 1.299 & \textbf{0.665} & 1.246 & 0.672 & \textbf{1.228} & 0.668 \\
        & & 36 & 1.241 & \textbf{0.676} & 1.242 & 0.678 & \textbf{1.239} & \textbf{0.676} & 1.320 & 0.743 & 1.346 & 0.748 \\
        & & 48 & 1.530 & 0.750 & 1.530 & \textbf{0.749} & \textbf{1.528} & \textbf{0.749} & 1.549 & 0.809 & 1.573 & 0.814 \\
        & & 60 & 1.406 & \textbf{0.731} & \textbf{1.404} & \textbf{0.731} & 1.406 & \textbf{0.731} & 1.525 & 0.829 & 1.630 & 0.849 \\
        \midrule
        & \multirow{4}{*}{{ETTh1}} & 96 & \textbf{0.360} & \textbf{0.386} & \textbf{0.360} & 0.387 & 0.366 & 0.387 & 0.365 & 0.385 & 0.375 & 0.398 \\
        & & 192 & \textbf{0.402} & \textbf{0.412} & 0.403 & 0.413 & 0.414 & 0.413 & 0.413 & \textbf{0.412} & 0.417 & 0.422 \\
        & & 336 & \textbf{0.414} & \textbf{0.421} & 0.416 & 0.426 & 0.424 & 0.426 & 0.426 & 0.423 & 0.422 & 0.431 \\
        & & 720 & 0.420 & 0.446 & 0.446 & 0.446 & 0.449 & \textbf{0.412} & \textbf{0.412} & 0.439 & 0.484 & 0.486 \\
       \midrule
        & \multirow{4}{*}{{ETTh2}} & 96 & \textbf{0.274} & 0.328 & \textbf{0.274} & 0.329 & 0.275 & 0.328 & 0.276 & 0.327 & 0.281 & 0.337 \\
        & & 192 & 0.339 & 0.371 & 0.338 & 0.372 & 0.340 & 0.371 & 0.341 & 0.370 & 0.355 & 0.387 \\
        & & 336 & \textbf{0.330} & \textbf{0.373} & 0.331 & 0.374 & 0.335 & 0.376 & 0.332 & \textbf{0.373} & 0.348 & 0.393 \\
        & & 720 & 0.382 & 0.415 & 0.383 & 0.417 & 0.386 & 0.416 & \textbf{0.378} & \textbf{0.406} & 0.396 & 0.432 \\
        \midrule
        & \multirow{4}{*}{{ETTm1}} & 96 & \textbf{0.283} & \textbf{0.328} & 0.284 & 0.330 & 0.289 & 0.328 & 0.450 & 0.362 & 0.293 & 0.345 \\
        & & 192 & \textbf{0.324} & \textbf{0.355} & \textbf{0.324} & 0.356 & 0.328 & \textbf{0.355} & 0.517 & 0.393 & 0.333 & 0.371 \\
        & & 336 & \textbf{0.356} & \textbf{0.376} & 0.357 & 0.377 & 0.364 & 0.377 & 0.558 & 0.414 & 0.363 & 0.389 \\
        & & 720 & 0.424 & 0.420 & \textbf{0.421} & 0.422 & 0.428 & \textbf{0.412} & 0.587 & 0.442 & 0.424 & 0.424 \\
        \midrule
        & \multirow{4}{*}{{ETTm2}} & 96 & 0.163 & 0.246 & \textbf{0.162} & 0.246 & 0.163 & \textbf{0.244} & 0.167 & 0.247 & 0.164 & 0.253 \\
        & & 192 & 0.216 & \textbf{0.283} & \textbf{0.215} & \textbf{0.283} & 0.220 & 0.284 & 0.222 & 0.285 & 0.222 & 0.292 \\
        & & 336 & \textbf{0.264} & \textbf{0.315} & \textbf{0.264} & \textbf{0.315} & 0.269 & 0.317 & 0.268 & 0.316 & 0.270 & 0.324 \\
        & & 720 & 0.348 & \textbf{0.370} & \textbf{0.347} & \textbf{0.370} & 0.353 & 0.371 & 0.353 & 0.371 & 0.357 & 0.381 \\
        \midrule
        &\multicolumn{2}{c|}{Average} & \textbf{0.539} & \textbf{0.433} & 0.540 & 0.435 & 0.543 & 0.434 & 0.585 & 0.451 & 0.564 & 0.457 \\
        \bottomrule
    \end{tabular}
}\vspace{-1mm}
    \caption{The ablation study on the smooth quadratic loss. The symbol ``-'' denotes the deletion. TimeSQL-SQL stands for TimeSQL using the MSE loss.}
    \vspace{-3mm}
    \label{tab:abs}
\end{table}

\begin{figure}[t]
    \centering
    \includegraphics[width=0.32\textwidth]{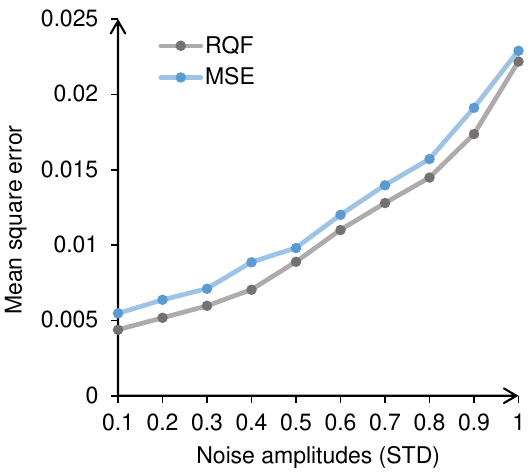}
    \vspace{-2.5mm}
    \caption{The simulation experiment with different amplitudes of Gaussian noises.}\label{fig:simulation}
    \vspace{-4.5mm}
\end{figure}


\subsection{Implementation Details}

For all datasets except for ILI, the input sequence length in TimeSQL is set to $336$, while the prediction length $T$ is chosen from the set $\{96, 192, 336, 720\}$. Since the sample number of the ILI dataset is relatively small, we follow \citeauthor{Yuqietal-2023-PatchTST}, \citeyear{Yuqietal-2023-PatchTST} to set the input sequence length to $104$ and select the prediction length $T$  from the pool of $\{24, 36, 48, 60\}$. The hyperparameters $c$, $\alpha$, $\beta$, and $\gamma$ in SQL for the other 7 datasets are manually set to 0.08, 0.2, 0.05, and 0.05, respectively.  The Adam algorithm with a learning rate of $1\times 10^{-4}$ is used to update the parameters. Due to the substantial difference in the ILI dataset, the hyperparameters are re-adjusted (See Appendix C). For all datasets,  we adopt LSTM as the temporal encoder of TimeSQL and apply reversible instance
normalization \cite{kim2021reversible} to stabilize the learning process.

\subsection{Simulation Results}
To validate the theoretical conclusion on the rational quadratic function (RQF) in TimeSQL, we present a simulation experiment on mock data. We conduct such simulation experiments because it is
hard to access clean data (no noise) with real data. Specifically, we build a TSF dataset by mixing the Gaussian noises into the trigonometric functions, each variable corresponding to a distinct function. We observe that, at each experiment with different amplitudes of the noises (indicated by the standard derivations), the performance of MLP with RQF loss is always better than that with MSE loss (Figure \ref{fig:simulation} and Appendix D).


\subsection{Main Results}
To assess the performance of TimeSQL, we apply TimeSQL to the 8 benchmark datasets and adopt the mean square error (MSE) and the mean absolute error (MAE) as the evaluation metrics. Table~\ref{tab:multivariate} reports the prediction performance of all the competitive baselines with different prediction lengths. The transformer-based architectures Informer, Pyraformer, and LogTrans obtain the highest mean square error (MSE) and mean absolute error (MAE), indicating the poorest forecasting results. MICN extracts the information from multiple scales of time series, yielding better performance. PatchTST leverages patching operation and employs transformer architecture, standing as the previous SOTA method. However, the above methods reckon without the data noises, leading to limited forecasting power. TimeSQL not only integrates the advanced techniques of the above models (multi-scale operation and patching) but also uses a new smooth quadratic loss, outperforming all the other methods with significant improvements on most of the datasets.

Specifically, TimeSQL surpasses the previously best method PatchTST on all the datasets in terms of MAE, achieving decreases of MAE by $5.3\%$, $5.7\%$, $1.2\%$, $13.4\%$, $4.1\%$, $2.9\%$, $3.1\%$, and $3.8\%$ for the Weather, Traffic, Electricity, ILI, ETTh1, ETTh2, ETTm1, and ETTm2 datasets, respectively. Besides, TimeSQL also performs significantly better in terms of the MSE metric on various datasets and prediction lengths. The above comparison of 8 benchmark datasets establishes that TimeSQL is a simple and effective time series forecasting method.

\subsection{Ablation Study}
We next investigate why TimeSQL presents an impressive forecasting performance with such a simple approach. 

To demonstrate the effectiveness of the multi-scale operation, we create a control model with only a single scale of patching (denoted as TimeSQL+SP). We test it on five benchmark datasets that have fewer features for fast evaluation. As shown in Table~\ref{tab:TimeSQL} the performance of TimeSQL+SP is reduced by a noticeable margin, revealing the importance of the multi-scale operation in TimeSQL. To validate the universality of the multi-scale patching, we take Transformer as the backbone in TimeSQL and further probe the role of multi-scale patching. We observe that TimeSQL with the multi-scale patching transformer (denoted as TimeSQL+MPT) is also better than the TimeSQL variant with a single-scale patching transformer (TimeSQL+SPT). The above results demonstrate the indispensable role of multi-scale patching in TimeSQL and other methods of time series forecasting.

\begin{table}[t]
	\centering
 \renewcommand{\arraystretch}{1.15}
	\resizebox{1.0\linewidth}{!}{
		\begin{tabular}{cc|c|cccc|ccccc}
			\toprule
			&\multicolumn{2}{c}{Methods}& \multicolumn{2}{c}{PatchTST+MSE}& \multicolumn{2}{c}{PatchTST+SQL}& \multicolumn{2}{c}{Informer+MSE}& \multicolumn{2}{c}{Informer+SQL}\\
			\cmidrule(r){4-5}\cmidrule(r){6-7}\cmidrule(r){8-9}\cmidrule(r){10-11}
			&\multicolumn{2}{c|}{Metric}&MSE&MAE&MSE&\multicolumn{1}{c}{MAE}&MSE&MAE&MSE&MAE\\
			\midrule
			&\multirow{4}*{{ILI}}& 24    & 1.522 & 0.814 & \textbf{1.412} & \textbf{0.722} & 4.657 & 1.449 & \textbf{4.603} & \textbf{1.406} \\
            &\multicolumn{1}{c|}{}& 36   & 1.430 & 0.834 & \textbf{1.351} & \textbf{0.714} & 4.650 & 1.463 & \textbf{4.614} & \textbf{1.430} \\
            &\multicolumn{1}{c|}{}& 48   & 1.673 & 0.854 & \textbf{1.634} & \textbf{0.789} & \textbf{5.004} & \textbf{1.542} & 5.402 & 1.561 \\
            &\multicolumn{1}{c|}{}& 60   & 1.529 & 0.862 & \textbf{1.478} & \textbf{0.750} & \textbf{5.071} & \textbf{1.543} & 5.272 & 1.564 \\
            \midrule
			&\multirow{4}*{{ETTh1}}& 96    & 0.375 & 0.399 & \textbf{0.360} & \textbf{0.387} & 0.941 & 0.769 & \textbf{0.846} & \textbf{0.680} \\
            &\multicolumn{1}{c|}{}& 192   & 0.414 & 0.421 & \textbf{0.407} & \textbf{0.418} & 1.007 & 0.786 & \textbf{0.976} & \textbf{0.723} \\
            &\multicolumn{1}{c|}{}& 336   & 0.431 & 0.436 & \textbf{0.422} & \textbf{0.426} & 1.038 & 0.784 & \textbf{1.027} & \textbf{0.762} \\
            &\multicolumn{1}{c|}{}& 720   & 0.449 & 0.466 & \textbf{0.427} & \textbf{0.452} & 1.144 & 0.857 & \textbf{1.136} & \textbf{0.817} \\
            \midrule
			&\multirow{4}*{{ETTh2}}& 96    & 0.274 & 0.336 & \textbf{0.273} & \textbf{0.329} & \textbf{1.549} & \textbf{0.952} & 1.779 & 1.004 \\
            &\multicolumn{1}{c|}{}& 192   & \textbf{0.339} & 0.379 & 0.341 & \textbf{0.373} & 3.792 & 1.542 & \textbf{1.878} & \textbf{1.042} \\
            &\multicolumn{1}{c|}{}& 336   & \textbf{0.331} & 0.380 & \textbf{0.331} & \textbf{0.376} & 4.215 & 1.642 & \textbf{2.115} & \textbf{1.097} \\
            &\multicolumn{1}{c|}{}& 720   & 0.379 & 0.422 & \textbf{0.378} & \textbf{0.415} & 3.656 & 1.619 & \textbf{2.057} & \textbf{1.111} \\
			\midrule
			&\multirow{4}*{{ETTm1}}& 96    & 0.290 & 0.342 & \textbf{0.287} & \textbf{0.329} & 0.626 & 0.560 & \textbf{0.524} & \textbf{0.482} \\
            &\multicolumn{1}{c|}{}& 192   & 0.332 & 0.369 & \textbf{0.331} & \textbf{0.357} & 0.725 & 0.619 & \textbf{0.692} & \textbf{0.565} \\
            &\multicolumn{1}{c|}{}& 336   & \textbf{0.366} & 0.392 & 0.371 & \textbf{0.380} & 1.005 & 0.741 & \textbf{0.818} & \textbf{0.647} \\
            &\multicolumn{1}{c|}{}& 720   & \textbf{0.420} & 0.424 & 0.424 & \textbf{0.414} & 1.133 & 0.845 & \textbf{1.108} & \textbf{0.800} \\
            \midrule
			&\multirow{4}*{{ETTm2}}& 96    & 0.165 & 0.255 & \textbf{0.161} & \textbf{0.246} & 0.355 & 0.462 & \textbf{0.332} & \textbf{0.402} \\
            &\multicolumn{1}{c|}{}& 192   & 0.220 & 0.292 & \textbf{0.219} & \textbf{0.286} & \textbf{0.595} & \textbf{0.586} & 0.816 & 0.654 \\
            &\multicolumn{1}{c|}{}& 336   & 0.278 & 0.329 & \textbf{0.269} & \textbf{0.320} & 1.270 & 0.871 & \textbf{1.176} & \textbf{0.844} \\
            &\multicolumn{1}{c|}{}& 720   & 0.367 & 0.385 & \textbf{0.348} & \textbf{0.371} & 3.001 & 1.267 & \textbf{2.224} & \textbf{1.141} \\
			\midrule
            &\multicolumn{2}{c|}{Average} & 0.579 & 0.470 & \textbf{0.561} & \textbf{0.443} & 2.272 & 1.075 & \textbf{1.970} & \textbf{0.937}\\
            \bottomrule
		\end{tabular}
	}\vspace{-1mm}
	\caption{The comparison between SQL and MSE.}
 \vspace{-4mm}
	\label{tab:othermodel}
\end{table}

We further build several TimeSQL variants by removing the sub-modules in the smooth quadratic loss (SQL), including TimeSQL without rational quadratic function (TimeSQL-RQF), the one without the outlier regularization (TimeSQL-OR), the one without MAE (TimeSQL-MAE), and the one without SQL (TimeSQL-SQL, where the MSE loss is used). Table~\ref{tab:abs} shows that deleting any sub-modules in the smooth quadratic loss would lead to the increase of MSE and MAE, indicating the usefulness of these sub-modules. Also, we see that the TimeSQL model with MSE loss presents the poorest performance among these variants, showing the effectiveness of the smooth quadratic loss. 

Besides, we also applied SQL to PatchTST and Informer, which are two widely-used TSF models. Table~\ref{tab:othermodel}  shows that these models with SQL perform largely better than the original ones, further demonstrating that SQL is a general and effective loss function for time series forecasting. See Appendix E for more ablation studies.
\subsection{Case Study}
We conduct a case study to showcase the forecasting behavior of our model on the electricity dataset. For representation simplicity, we only depict the changes of one variable. In Figure \ref{fig:case}, we observe that, based on the input with the length 336, both TimeSQL and PatchTST could apprehend the fundamental temporal dynamics; The prediction curve of TimeSQL is more consistent with the ground truth than that of PatchTST. Next, we manually add the Gaussian noises into the input of this test sample. The models could only receive the input mixed with noises and are required to predict the future time series. Due to the influences of the noises, the predictions of both models deviate from the ground truth more than the original cases. However, we notice that the deviation of TimeSQL is not large, and even smaller than the deviation of PatchTST in the original sample. This comparison further illustrates the superiority of TimeSQL in learning temporal dynamics underlying the noising time series. See Appendix F for the error cases of TimeSQL.

 \begin{figure}[t]
    \centering
    \includegraphics[width=0.48\textwidth]{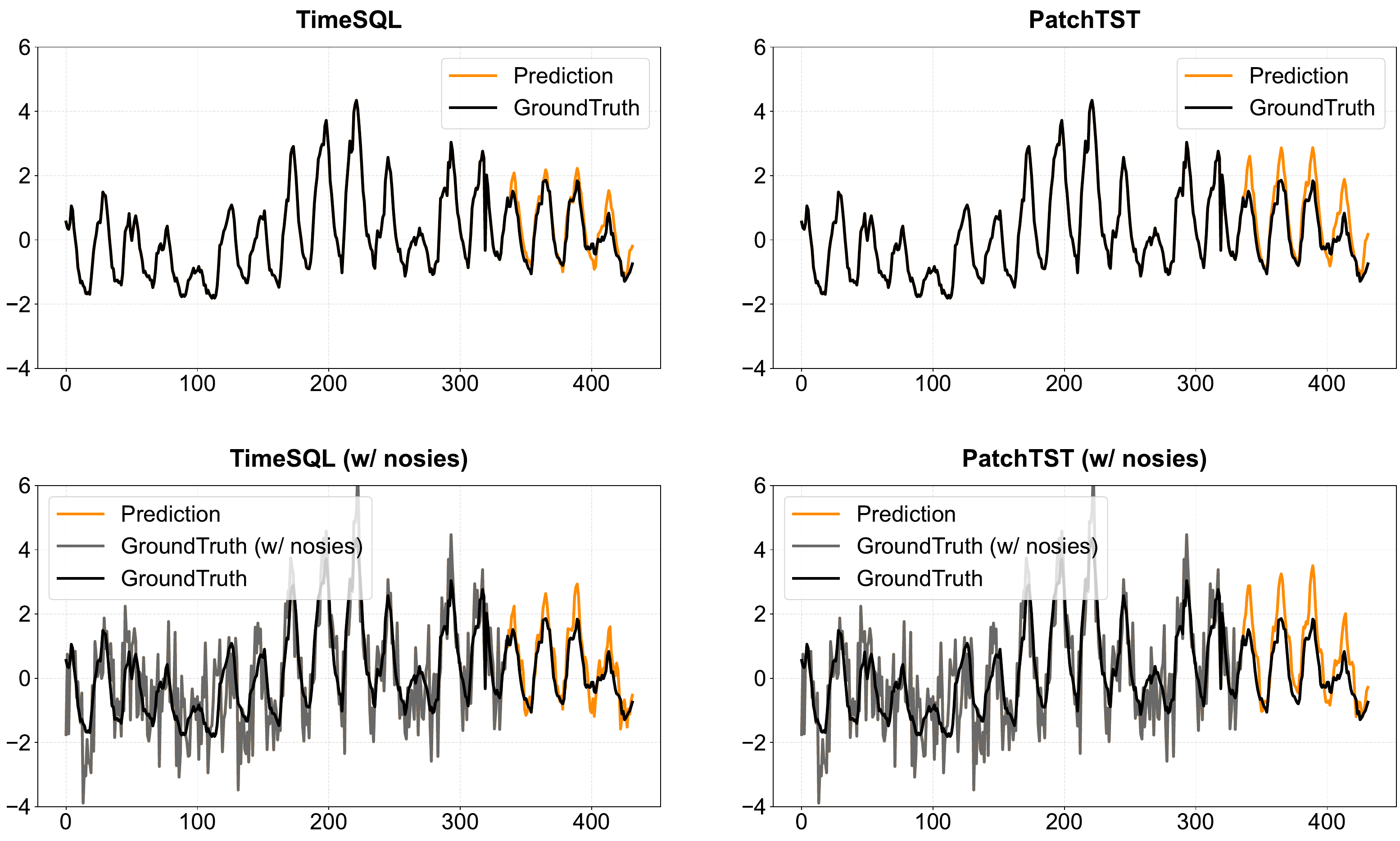}\vspace{-2mm}
    \caption{Case study of TimeSQL and PatchTST.}\label{fig:case}
    \vspace{-3mm}
\end{figure}

\section{Conclusion and Discussion}
This paper introduces a simple but effective method, coined as TimeSQL, for multivariate time series forecasting. In TimeSQL, we propose multi-scale patching and smooth quadratic loss (SQL) to learn the essential long-term dynamics of the time series. Theoretical analysis shows that, under certain simple conditions, the effect of the noises on the model with smooth quadratic loss is smaller than the model with MSE. 
The limitation of TimeSQL mainly lies in the absence of the scope of using SQL and the lack of dynamics analysis in the model learning. We will further develop related theorems in the future.

\bibliography{aaai24}

\begin{thebibliography}{29}
\providecommand{\natexlab}[1]{#1}

\bibitem[{Bai, Kolter, and Koltun(2018)}]{bai2018empirical}
Bai, S.; Kolter, J.~Z.; and Koltun, V. 2018.
\newblock An empirical evaluation of generic convolutional and recurrent networks for sequence modeling.
\newblock \emph{arXiv preprint arXiv:1803.01271}.

\bibitem[{Bartholomew(1971)}]{bartholomew1971time}
Bartholomew, D.~J. 1971.
\newblock Time Series Analysis Forecasting and Control.

\bibitem[{Chen et~al.(2015)Chen, He, Benesty, Khotilovich, Tang, Cho, Chen, Mitchell, Cano, Zhou et~al.}]{chen2015xgboost}
Chen, T.; He, T.; Benesty, M.; Khotilovich, V.; Tang, Y.; Cho, H.; Chen, K.; Mitchell, R.; Cano, I.; Zhou, T.; et~al. 2015.
\newblock Xgboost: extreme gradient boosting.
\newblock \emph{R package version 0.4-2}, 1(4): 1--4.

\bibitem[{Cuturi and Blondel(2017)}]{cuturi2017soft}
Cuturi, M.; and Blondel, M. 2017.
\newblock Soft-dtw: a differentiable loss function for time-series.
\newblock In \emph{International conference on machine learning}, 894--903. PMLR.

\bibitem[{Dey and Salem(2017)}]{dey2017gate}
Dey, R.; and Salem, F.~M. 2017.
\newblock Gate-variants of gated recurrent unit (GRU) neural networks.
\newblock In \emph{2017 IEEE 60th international midwest symposium on circuits and systems (MWSCAS)}, 1597--1600. IEEE.

\bibitem[{Dosovitskiy et~al.(2020)Dosovitskiy, Beyer, Kolesnikov, Weissenborn, Zhai, Unterthiner, Dehghani, Minderer, Heigold, Gelly et~al.}]{dosovitskiy2020image}
Dosovitskiy, A.; Beyer, L.; Kolesnikov, A.; Weissenborn, D.; Zhai, X.; Unterthiner, T.; Dehghani, M.; Minderer, M.; Heigold, G.; Gelly, S.; et~al. 2020.
\newblock An image is worth 16x16 words: Transformers for image recognition at scale.
\newblock \emph{arXiv preprint arXiv:2010.11929}.

\bibitem[{Graves and Graves(2012)}]{graves2012long}
Graves, A.; and Graves, A. 2012.
\newblock Long short-term memory.
\newblock \emph{Supervised sequence labelling with recurrent neural networks}, 37--45.

\bibitem[{Hearst et~al.(1998)Hearst, Dumais, Osuna, Platt, and Scholkopf}]{hearst1998support}
Hearst, M.~A.; Dumais, S.~T.; Osuna, E.; Platt, J.; and Scholkopf, B. 1998.
\newblock Support vector machines.
\newblock \emph{IEEE Intelligent systems and their applications}, 13(4): 18--28.

\bibitem[{Kadiyala and Kumar(2014)}]{kadiyala2014multivariate}
Kadiyala, A.; and Kumar, A. 2014.
\newblock Multivariate time series models for prediction of air quality inside a public transportation bus using available software.
\newblock \emph{Environmental progress \& Sustainable energy}, 33(2): 337--341.

\bibitem[{Kardakos et~al.(2013)Kardakos, Alexiadis, Vagropoulos, Simoglou, Biskas, and Bakirtzis}]{kardakos2013application}
Kardakos, E.~G.; Alexiadis, M.~C.; Vagropoulos, S.~I.; Simoglou, C.~K.; Biskas, P.~N.; and Bakirtzis, A.~G. 2013.
\newblock Application of time series and artificial neural network models in short-term forecasting of PV power generation.
\newblock In \emph{2013 48th International Universities' Power Engineering Conference (UPEC)}, 1--6. IEEE.

\bibitem[{Kim et~al.(2021)Kim, Kim, Tae, Park, Choi, and Choo}]{kim2021reversible}
Kim, T.; Kim, J.; Tae, Y.; Park, C.; Choi, J.-H.; and Choo, J. 2021.
\newblock Reversible instance normalization for accurate time-series forecasting against distribution shift.
\newblock In \emph{International conference on learning representations}.

\bibitem[{Lai et~al.(2018)Lai, Chang, Yang, and Liu}]{lai2018modeling}
Lai, G.; Chang, W.-C.; Yang, Y.; and Liu, H. 2018.
\newblock Modeling long-and short-term temporal patterns with deep neural networks.
\newblock In \emph{The 41st international ACM SIGIR conference on research \& development in information retrieval}, 95--104.

\bibitem[{Le~Guen and Thome(2019)}]{le2019shape}
Le~Guen, V.; and Thome, N. 2019.
\newblock Shape and time distortion loss for training deep time series forecasting models.
\newblock \emph{Advances in neural information processing systems}, 32.

\bibitem[{Li et~al.(2019)Li, Jin, Xuan, Zhou, Chen, Wang, and Yan}]{li2019enhancing}
Li, S.; Jin, X.; Xuan, Y.; Zhou, X.; Chen, W.; Wang, Y.-X.; and Yan, X. 2019.
\newblock Enhancing the locality and breaking the memory bottleneck of transformer on time series forecasting.
\newblock \emph{Advances in neural information processing systems}, 32.

\bibitem[{Liu et~al.(2021)Liu, Yu, Liao, Li, Lin, Liu, and Dustdar}]{liu2021pyraformer}
Liu, S.; Yu, H.; Liao, C.; Li, J.; Lin, W.; Liu, A.~X.; and Dustdar, S. 2021.
\newblock Pyraformer: Low-complexity pyramidal attention for long-range time series modeling and forecasting.
\newblock In \emph{International conference on learning representations}.

\bibitem[{Maclaurin(1742)}]{maclaurin1742treatise}
Maclaurin, C. 1742.
\newblock \emph{A treatise of fluxions: in two books}, volume~1.
\newblock Ruddimans.

\bibitem[{McCloskey and Cohen(1989)}]{mccloskey1989catastrophic}
McCloskey, M.; and Cohen, N.~J. 1989.
\newblock Catastrophic interference in connectionist networks: The sequential learning problem.
\newblock In \emph{Psychology of learning and motivation}, volume~24, 109--165. Elsevier.

\bibitem[{Medsker and Jain(2001)}]{medsker2001recurrent}
Medsker, L.~R.; and Jain, L. 2001.
\newblock Recurrent neural networks.
\newblock \emph{Design and applications}, 5(64-67): 2.

\bibitem[{Morid, Sheng, and Dunbar(2023)}]{morid2023time}
Morid, M.~A.; Sheng, O. R.~L.; and Dunbar, J. 2023.
\newblock Time series prediction using deep learning methods in healthcare.
\newblock \emph{ACM Transactions on management information systems}, 14(1): 1--29.

\bibitem[{Nie et~al.(2023)Nie, H.~Nguyen, Sinthong, and Kalagnanam}]{Yuqietal-2023-PatchTST}
Nie, Y.; H.~Nguyen, N.; Sinthong, P.; and Kalagnanam, J. 2023.
\newblock A Time Series is Worth 64 Words: Long-term Forecasting with Transformers.
\newblock In \emph{International conference on learning representations}.

\bibitem[{Radford et~al.(2018)Radford, Narasimhan, Salimans, Sutskever et~al.}]{radford2018improving}
Radford, A.; Narasimhan, K.; Salimans, T.; Sutskever, I.; et~al. 2018.
\newblock Improving language understanding by generative pre-training.

\bibitem[{Vaswani et~al.(2017)Vaswani, Shazeer, Parmar, Uszkoreit, Jones, Gomez, Kaiser, and Polosukhin}]{vaswani2017attention}
Vaswani, A.; Shazeer, N.; Parmar, N.; Uszkoreit, J.; Jones, L.; Gomez, A.~N.; Kaiser, {\L}.; and Polosukhin, I. 2017.
\newblock Attention is all you need.
\newblock \emph{Advances in neural information processing systems}, 30.

\bibitem[{Wang et~al.(2023)Wang, Peng, Huang, Wang, Chen, and Xiao}]{micn}
Wang, H.; Peng, J.; Huang, F.; Wang, J.; Chen, J.; and Xiao, Y. 2023.
\newblock MICN: Multi-scale Local and Global Context Modeling for Long-term Series Forecasting.

\bibitem[{Wu et~al.(2023)Wu, Hu, Liu, Zhou, Wang, and Long}]{wu2023timesnet}
Wu, H.; Hu, T.; Liu, Y.; Zhou, H.; Wang, J.; and Long, M. 2023.
\newblock TimesNet: Temporal 2D-Variation Modeling for General Time Series Analysis.
\newblock In \emph{International conference on learning representations}.

\bibitem[{Wu et~al.(2021)Wu, Xu, Wang, and Long}]{wu2021autoformer}
Wu, H.; Xu, J.; Wang, J.; and Long, M. 2021.
\newblock Autoformer: Decomposition transformers with auto-correlation for long-term series forecasting.
\newblock \emph{Advances in neural information processing systems}, 34: 22419--22430.

\bibitem[{Zeng et~al.(2023)Zeng, Chen, Zhang, and Xu}]{zeng2023transformers}
Zeng, A.; Chen, M.; Zhang, L.; and Xu, Q. 2023.
\newblock Are transformers effective for time series forecasting?
\newblock In \emph{Proceedings of the AAAI conference on artificial intelligence}, volume~37, 11121--11128.

\bibitem[{Zhou et~al.(2021)Zhou, Zhang, Peng, Zhang, Li, Xiong, and Zhang}]{zhou2021informer}
Zhou, H.; Zhang, S.; Peng, J.; Zhang, S.; Li, J.; Xiong, H.; and Zhang, W. 2021.
\newblock Informer: Beyond efficient transformer for long sequence time-series forecasting.
\newblock In \emph{Proceedings of the AAAI conference on artificial intelligence}, volume~35, 11106--11115.

\bibitem[{Zhou et~al.(2022)Zhou, Ma, Wen, Wang, Sun, and Jin}]{zhou2022fedformer}
Zhou, T.; Ma, Z.; Wen, Q.; Wang, X.; Sun, L.; and Jin, R. 2022.
\newblock Fedformer: Frequency enhanced decomposed transformer for long-term series forecasting.
\newblock In \emph{International conference on machine learning}, 27268--27286. PMLR.

\bibitem[{Zhu, Ma, and Lin(2019)}]{zhu2019detecting}
Zhu, Q.; Ma, H.; and Lin, W. 2019.
\newblock Detecting unstable periodic orbits based only on time series: When adaptive delayed feedback control meets reservoir computing.
\newblock \emph{Chaos: An Interdisciplinary Journal of Nonlinear Science}, 29(9).

\end{thebibliography}

\appendices
\section{Proof of Theorem 1}

\noindent \emph{Proof. } For a prediction $\hat{x}$ and the ground truth $y$ of a particular time step and variable, the MSE loss is:
\begin{align}
Loss_{\text{MSE}}=(y-\hat{x})^2=e^2, 
\end{align}
where $e$ represents the error between the ground truth and the predicted value.

The training loss with noise $\varepsilon$ is:
\begin{equation}
\begin{aligned}
Loss^{'}_{\text{MSE}}=(y+\varepsilon-\hat{x})^2=(e+\varepsilon)^2 \\
\end{aligned}
\end{equation}
The RQF loss is: 
\begin{equation}
\begin{aligned}
Loss_{\text{RQF}}=\frac{e^2}{e^2+c} \\
\end{aligned}
\end{equation}
where $c\in (0,+\infty)$. The RQF loss with noise $\varepsilon$ is:
\begin{equation}
\begin{aligned}
Loss^{'}_{\text{RQF}}=\frac{(e+\varepsilon )^2}{(e+\varepsilon )^2+c} \\
\end{aligned}
\end{equation}
According to definition 1, the effect of the noise on the MSE loss is measured by
\begin{equation}
\begin{aligned}
V_{\text{MSE}}=\frac{\left | Loss^{'}_{\text{MSE}}-Loss_{\text{MSE}} \right | }{Loss_{\text{MSE}}} =\frac{\left | 2\varepsilon e+\varepsilon ^2 \right |}{e^2} \\
\end{aligned}
\end{equation}

Similarly, the effect of the noise on the RQF loss is measured by
\begin{equation}
\begin{aligned}
V_{\text{RQF}}=\frac{\left | Loss^{'}_{\text{RQF}}-Loss_{\text{RQF}} \right | }{Loss_{\text{RQF}}} =\frac{c\left | 2\varepsilon e+\varepsilon ^2 \right |}{e^2[(e+\varepsilon )^2+c]}  \\
\end{aligned}
\end{equation}
The comparison between RQF and MSE in terms of the amplitude of the loss changes can be measured by
\begin{equation}
\begin{aligned}
\frac{V_{\text{RQF}}}{V_{\text{MSE}}}=\frac{c}{c+(e+\varepsilon )^2}\le   1
\end{aligned}
\end{equation}
Considering $c\in (0,+\infty)$, the value of the above division is smaller than 1. Thus, we have
\begin{equation}
V_{\text{RQF}} < V_{\text{MSE}}
\end{equation}

Therefore, we have the following conclusion: regardless of the distribution of the noise $\varepsilon$ in the time-series data, the effect of the noise on the RQF loss is always lower than that of MSE loss.\ \ \ \ \ \  \ \ \ \ \ \ \ \ \ \ \ \ \  \ \ \ \ \ \ \ \ \ \ \ \ \  \ \ \ \ \ \ \ \ \ \ \ \ \  \ \ \ \ \ \ \ \ \ \ \ \ \  \ \ \ \ \ \ \ \ \ \ \ \ \  \ \ \ \ \ \ \  $\square$

\section{Proof of Theorem 2}

\textit{Proof.} We first simplify the above formulas.
\begin{align}
\nabla_\theta  \rm RQF(y,\hat{x})&=\frac{\partial \rm RQF(y,\hat{x})}{\partial \theta} \\
&=\frac{\partial \rm RQF(y,\hat{x})}{\partial \hat{x}} \frac{\partial \hat{x}}{\partial \theta} \\
&= \frac{2c(\hat{x}-y)}{((\hat{x}-y)^2 + c)^2}\frac{\partial \hat{x}}{\partial \theta}\\
&= \frac{2ce}{(e^2 + c)^2}\frac{\partial \hat{x}}{\partial \theta}.
\end{align}
Similarly, we have
\begin{align}
\nabla_\theta  \rm RQF(y+\varepsilon,\hat{x})&= \frac{2c(e+\varepsilon)}{((e+\varepsilon)^2 + c)^2}\frac{\partial \hat{x}}{\partial \theta}, \\
\nabla_\theta  \rm MSE(y,\hat{x}) &= 2e \frac{\partial \hat{x}}{\partial \theta},\\
\nabla_\theta  \rm MSE(y+\varepsilon,\hat{x})&=2(e+\varepsilon)\frac{\partial \hat{x}}{\partial \theta}.
\end{align}

Therefore, the left term  in Theorem 2 is given by
\begin{align}
V_r &= \left | \frac{\nabla_\theta \rm RQF(y+\varepsilon,\hat{x})- \nabla_\theta \rm RQF(y,\hat{x})}{\nabla_\theta  \rm RQF(y,\hat{x})}\right |\\
&= \left | \frac{ \frac{2c(e+\varepsilon)}{((e+\varepsilon)^2 + c)^2} -  \frac{2ce}{(e^2 + c)^2} }{\frac{2ce}{(e^2 + c)^2}} \right | \\
&= \left | \frac{ (e+\varepsilon) (e^2 + c)^2 - e ((e+\varepsilon)^2 + c)^2  } { e ((e+\varepsilon)^2 + c)^2 } \right |
\end{align}

The right term  in Theorem 2 is given by
\begin{align}
V_m &= \left | \frac{\nabla_\theta \rm MSE(y+\varepsilon,\hat{x})  - \nabla_\theta \rm MSE(y,\hat{x})}{\nabla_\theta \rm MSE(y,\hat{x})}\right |\\
&= \left | \frac{\varepsilon}{e}  \right |
\end{align}

The difference between $V_r$ and $V_m$ is
\begin{small}
\begin{align}
&V_r - V_m \nonumber\\
&= \frac{ \left | (e+\varepsilon) (e^2 + c)^2 - e ((e+\varepsilon)^2 + c)^2 \right |-\left | \varepsilon \right |  ((e+\varepsilon)^2 + c)^2} { |e| ((e+\varepsilon)^2 + c)^2 } 
\end{align}
\end{small}
As we aim to prove $V_r$ is smaller than $V_m$, we next only focus on whether the numerator of the above fraction is lower than 0.
    
\noindent$\bullet$ \textbf{Condition A.} If $\varepsilon \ge 2|e|$ and $((e+\varepsilon) (e^2 + c)^2 - e ((e+\varepsilon)^2 + c)^2)\ge 0$ , we have
\begin{small}
\begin{align}
C_a &= (e+\varepsilon) (e^2 + c)^2 - e ((e+\varepsilon)^2 + c)^2 -\varepsilon  ((e+\varepsilon)^2 + c)^2\\
&=-\varepsilon(e+\varepsilon)(2e+\varepsilon)[e^2+(e+\varepsilon)^2 +2c]\le 0
\end{align}
\end{small}
\noindent$\bullet$ \textbf{Condition B.} If $\varepsilon \ge 2|e|$ and $((e+\varepsilon) (e^2 + c)^2 - e ((e+\varepsilon)^2 + c)^2)<0$ , we have
\begin{small}
\begin{align}
C_b&= -(e+\varepsilon) (e^2 + c)^2 + e ((e+\varepsilon)^2 + c)^2 -\varepsilon  ((e+\varepsilon)^2 + c)^2\\
&=-(e+\varepsilon) (e^2 + c)^2 - (\varepsilon-e) ((e+\varepsilon)^2 + c)^2 \le 0
\end{align}
\end{small}

Based on the above analysis (Conditions A and B), we can conclude that, if $\varepsilon \ge 2|e|$, we have $V_r - V_m\le 0$.

\noindent$\bullet$ \textbf{Condition C.} If $\varepsilon \le -2|e|$ and $((e+\varepsilon) (e^2 + c)^2 - e ((e+\varepsilon)^2 + c)^2)\ge 0$ , we have
\begin{small}
\begin{align}
C_c&= (e+\varepsilon) (e^2 + c)^2 - e ((e+\varepsilon)^2 + c)^2 +\varepsilon  ((e+\varepsilon)^2 + c)^2\\
&=(e+\varepsilon) (e^2 + c)^2 + (\varepsilon-e) ((e+\varepsilon)^2 + c)^2 \le 0
\end{align}
\end{small}
\noindent$\bullet$ \textbf{Condition D.} If $\varepsilon \le -2|e|$ and $((e+\varepsilon) (e^2 + c)^2 - e ((e+\varepsilon)^2 + c)^2)<0$ , we have
\begin{small}
\begin{align}
C_d&= -(e+\varepsilon) (e^2 + c)^2 + e ((e+\varepsilon)^2 + c)^2 +\varepsilon  ((e+\varepsilon)^2 + c)^2\\
&=\varepsilon(e+\varepsilon)(2e+\varepsilon)[e^2+(e+\varepsilon)^2 +2c]\le 0
\end{align}
\end{small}
Based on the above analysis (Conditions A, B, C, and D), we can conclude that, if $|\varepsilon| \ge 2|e|$, we have $V_r - V_m\le 0$. Therefore, if $|\varepsilon|\ge 2|\hat{x}-y|$, we have
\begin{align}
&\!\!\!\!\!\!\!\!\!\!\!\left | \frac{\nabla_\theta \rm RQF(y+\varepsilon,\hat{x})- \nabla_\theta \rm RQF(y,\hat{x})}{\nabla_\theta  \rm RQF(y,\hat{x})}\right | \nonumber \\  & \le \left | \frac{\nabla_\theta \rm MSE(y+\varepsilon,\hat{x})  - \nabla_\theta \rm MSE(y,\hat{x})}{\nabla_\theta \rm MSE(y,\hat{x})}\right |
\end{align}
 $\square$

\textbf{Appendix Remarks 1.} Indeed, a tighter constraint exists for the potential range of the noise $\varepsilon$ (partially dependent on $c$).  However, since the analytic solution of the above inequality is unavailable, we leave it as future work to make a more exact solution about the boundary of the noise $\varepsilon$.

\section{Implementation Details}
\label{sec:appendix.med}
Appendix Table~\ref{tab:hptuning} shows the selection range of the hyperparameters of our model.
\begin{table}[!h]
\centering
\renewcommand{\arraystretch}{1.15}
\resizebox{\linewidth}{!}{
\begin{tabular}{c|c}
\toprule
Parameter & Range \\
\midrule
\texttt{BatchSize} &  [8, 16, 32, 128]\\
\hline
\texttt{HiddenSize} & [32, 64, 128, 256, 448] \\
\hline
\texttt{(Patch length1, Stride1)} & [(16,4),(16,8),(24,4),(34,2)] \\
\hline
\texttt{(Patch length2, Stride2)} & [(48,12),(48,24),(72,12),(68,4)] \\
\hline
\texttt{(Patch length3, Stride3)} & [(96,24),(96,48),(144,24),(102,12)] \\
\hline
\texttt{$\alpha$} & [0.1, 0.2] \\
\hline
\texttt{$c$} & [0.08, 100] \\
\hline
\texttt{$\beta$} & [0.0005, 0.05] \\
\hline
\texttt{$\gamma$} & [0.0001, 0.05] \\
\hline
\texttt{learningRate} &  [1e-4, 2.5e-4]\\
\hline
\texttt{revIn} & [True, False] \\
\bottomrule
\end{tabular}
}
\caption{Ranges of different hyper-paramaters}
\label{tab:hptuning}
\end{table}
The hyperparameters we select for each dataset are shown in Appendix table~\ref{tab:allhparams}.

\begin{table*}[htbp!]
  \centering
  \renewcommand{\arraystretch}{1.05}
  \resizebox{\linewidth}{!}{
  \begin{tabular}{c|c|c|c|c|c|c|c|c|c|c|c}
    \toprule
    Dataset & \texttt{BatchSize} & \texttt{HiddenSize} & \texttt{(Patch length1, Stride1)} & \texttt{(Patch length2, Stride2)} & \texttt{(Patch length3, Stride3)} & \texttt{$\alpha$} & \texttt{$c$} & \texttt{$\beta$} & \texttt{$\gamma$} & \texttt{learningRate} & \texttt{revIn} \\
    \midrule
    Weather & 128 & 128 & (16,8) & (48,24) & (96,48) & 0.2 & 0.08 & 0.05 & 0.05 & 1e-4 &True \\ \hline
    Traffic & 8 & 448 & (24,4) & (72,12) & (144,24) & 0.2 & 0.08 & 0.05 & 0.05 & 1e-4 &True \\ \hline
    Electricity & 32 & 256 & (16,4) & (48,12) & (96,24) & 0.2 & 0.08 & 0.05 & 0.05 & 1e-4 &True \\ \hline
    ILI & 16 & 32 & (34,2) & (68,4) & (102,12) & 0.1 & 100 & 0.0005 & 0.0001 & 2.5e-4 &True \\ \hline
    ETTh1 & 128 & 64 & (16,8) & (48,24) & (96,48) & 0.2 & 0.08 & 0.05 & 0.05 & 1e-4 &True \\ \hline
    ETTh2 & 128 & 64 & (16,8) & (48,24) & (96,48) & 0.2 & 0.08 & 0.05 & 0.05 & 1e-4 &True \\ \hline
    ETTm1 & 128 & 256 & (16,8) & (48,24) & (96,48) & 0.2 & 0.08 & 0.05 & 0.05 & 1e-4 &True \\ \hline
    ETTm2 & 128 & 256 & (16,8) & (48,24) & (96,48) & 0.2 & 0.08 & 0.05 & 0.05 & 1e-4 &True \\ 
    \bottomrule
  \end{tabular}
}  
  \caption{Model hyperparameter on each dataset.}\label{tab:allhparams}
\end{table*}
\begin{figure*}[htbp]
\includegraphics[width=1\textwidth,height=0.5\textwidth]{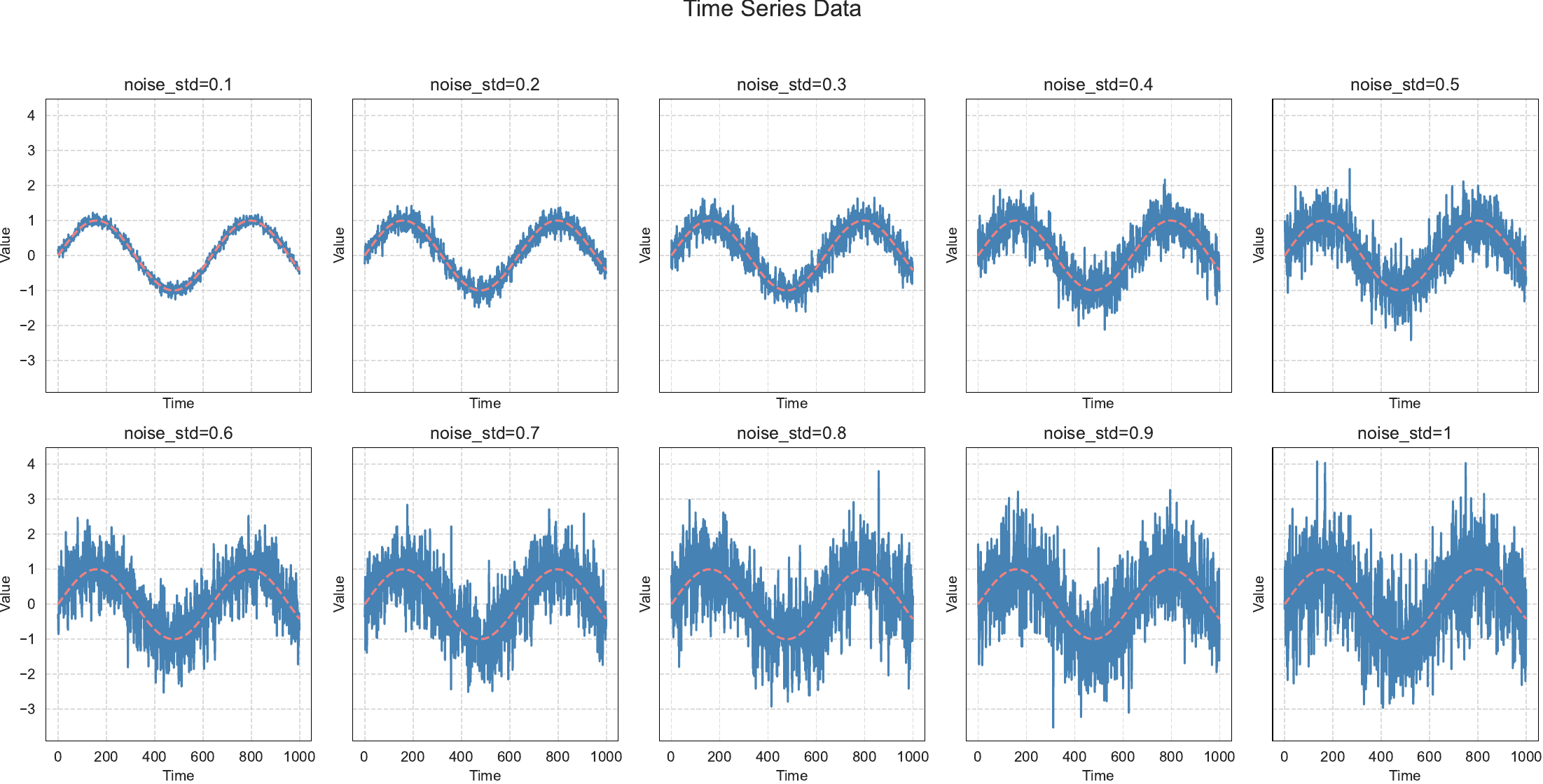}
\caption{Data with different amplitudes (std).}\label{fig:different std}
\end{figure*}

\begin{figure*}[htbp]
\includegraphics[width=1\textwidth,height=0.5\textwidth]{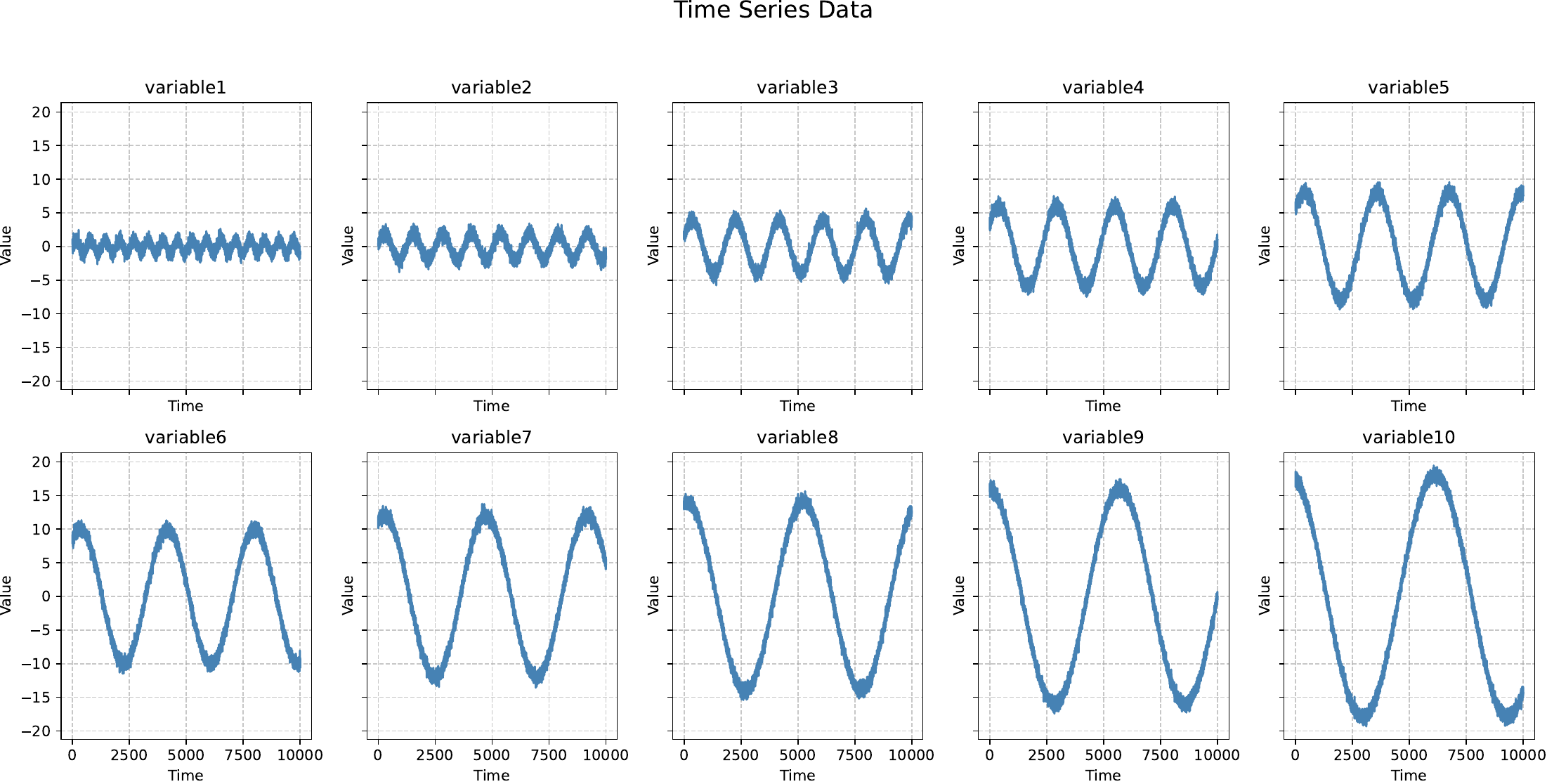}
\caption{Different variables in a dataset.}\label{fig:different variable}
\end{figure*}

\section{Simulation Experiment}
\label{sec.appendix.ane}
We construct several time series datasets for simulation experiments, each comprising $20,000$ data points and encompassing ten distinct variables. These variables are designed to adhere to distinct triangular functions characterized by a range of amplitudes, phases, and periods. For each task (i.e., a dataset), every variable in our dataset follows a unique triangular function. The amplitudes for these functions were selected from the set $amplitude \in \{1, 2, 4, 6, 8, 10, 12, 14, 16, 18\}$, phases from the set $phase \in \{0, 0.2, 0.4, 0.6, 0.8, 1, 1.2, 1.4, 1.6, 1.8\}$, and periods from the set $period \in \{1, 2, 3, 4, 5, 6, 7, 8, 9, 10\}$. The formula governing each variable adheres to its corresponding trigonometric function, ensuring diversity and complexity within the dataset.

\begin{equation} 
\begin{aligned}
Variable_1(t) &= \text{amplitude}_1 \cdot \sin\left(\frac{2\pi \cdot t}{\text{period}_1} + \text{phase}_1\right) \\
Variable_2(t) &= \text{amplitude}_2 \cdot \sin\left(\frac{2\pi \cdot t}{\text{period}_2} + \text{phase}_2\right) \\
&\vdots \\
Variable_{10}(t) &= \text{amplitude}_{10} \cdot \sin\left(\frac{2\pi \cdot t}{\text{period}_{10}} + \text{phase}_{10}\right)
\end{aligned}
\end{equation}

In order to mimic real-world scenarios, Gaussian noises are incorporated into the dataset. Specifically, various standard deviations of Gaussian noise are added to the first $80\%$ of data points within each dataset. Notably, the remaining $20\%$ of data points remained pristine, devoid of any artificially introduced noise. The partitioning of data into training and testing subsets was executed as follows: the initial $80\%$ of data points served as the training set, while the final $20\%$ constituted the test set. 

We introduce the Gaussian noises with a mean of zero and different standard deviations into the first 80\% of data for ten datasets, where $\text{std} \in \{0.1, 0.2, 0.3, 0.4, 0.5, 0.6, 0.7, 0.8, 0.9, 1\}$. The resulting dataset curves are illustrated in the  Appendix Figure~\ref{fig:different std}. The visual representation of all variables in one of the datasets is depicted in Appendix Figure~\ref{fig:different variable}. The experimental results are shown in Figure 3 of the main text.

\section{Ablation Studies on Prediction Lengths}
\label{sec:appendix.eas}
\begin{figure*}[t]
    \centering
    \includegraphics[width=0.95\textwidth]{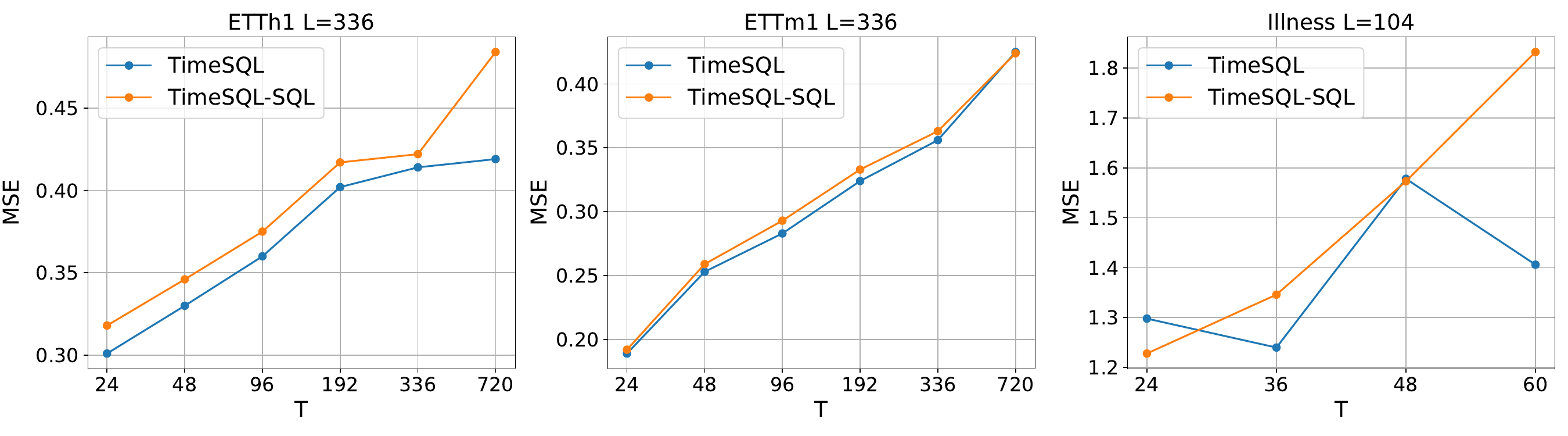}
    \caption{The comparison of individual prediction lengths in terms of MSE.}\label{fig:MSE}
\end{figure*}

\begin{figure*}[t]
    \centering
    \includegraphics[width=0.95\textwidth]{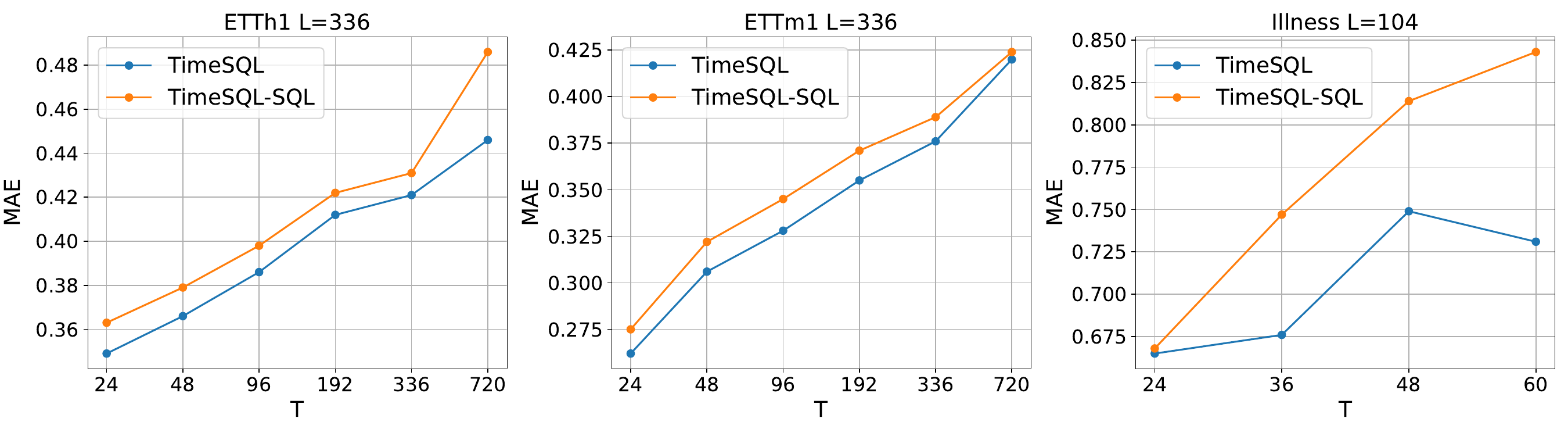}
    \caption{The comparison of individual prediction lengths in terms of MAE.} \label{fig:MAE}
\end{figure*}

In principle, the increase in prediction length leads to increased difficulty in prediction. Appendix Figures ~\ref{fig:MSE} and ~\ref{fig:MAE} depict the changes in MSE and MAE of TimeSQL-SQL and TimeSQL, respectively. TimeSQL-SQL stands for the TimeSQL model with the MSE loss.    
 The results indicate that  the prediction errors of TimeSQL increase slower  than those of TimeSQL-SQL along with the growing of the prediction length. The small errors of long prediction length are crucial for long-term prediction, highlighting the superiority of SQL in long-term prediction.

\section{Analysis on Error Cases}
Here, we select more samples where TimeSQL performs poorly for the deep analyses of the behaviors of our model. As depicted in Appendix Figure ~\ref{fig:case}, when the historical time series does not have an explicit period or pattern, the forecasting models (e.g., TimeSQL and PatchTST) struggle to accurately predict the future values. We also find that, even in these cases, the predictive curve of TimeSQL exhibits a higher degree of smoothness and is more consistent with the ground truth compared with the predictive curve of PatchTST.
 \begin{figure*}[t]
    \centering
    \includegraphics[width=0.8\textwidth]{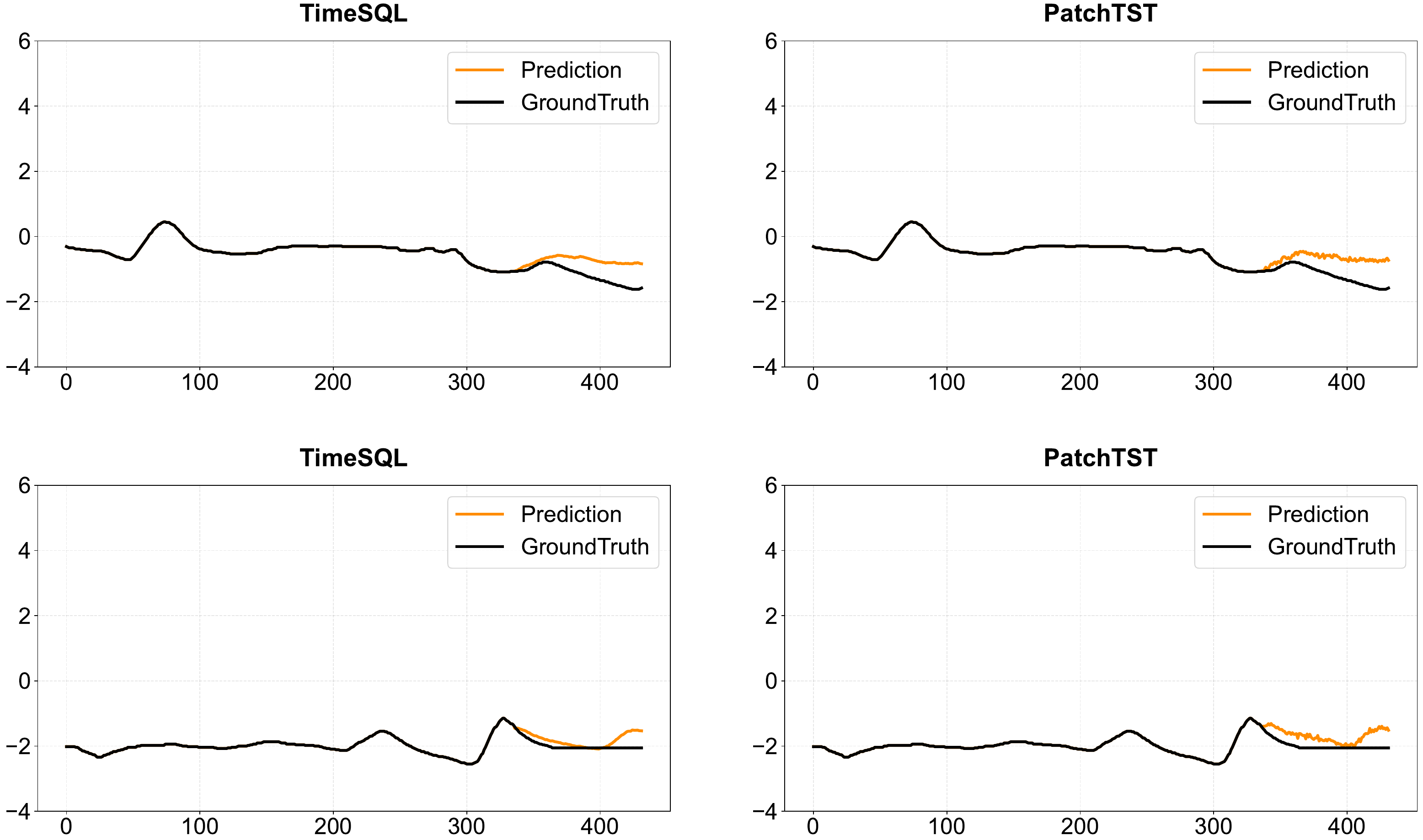}
    \caption{The predictions of TimeSQL and PatchTST on the difficult samples.}\label{fig:case}
\end{figure*}


\end{document}